\pdfoutput=1

\documentclass[11pt]{article}

\usepackage[]{ACL2023}

\usepackage{times}
\usepackage{latexsym}

\usepackage[T1]{fontenc}

\usepackage[utf8]{inputenc}

\usepackage{microtype}

\usepackage{inconsolata}
\usepackage{upgreek,textgreek}

\usepackage{booktabs}
\usepackage{colortbl}
\usepackage{multirow}
\usepackage[justification=Justified]{caption}
\usepackage{subcaption}
\usepackage{lscape}
\usepackage{graphicx}
\usepackage{xspace}
\usepackage{tcolorbox}
\usepackage{amsmath}

\newcommand{\SciArena}{\textsc{SciArena}\xspace}
\newcommand{\PubMed}{PubMed\xspace}
\newcommand{\PaperQA}{PaperQA2\xspace}
\newcommand{\SQ}{\textit{SQ}\xspace}
\newcommand{\MQ}{\textit{MQ}\xspace}
\newcommand{\MQCAT}{\textit{MQ}$_{+Cat}$\xspace}
\newcommand{\MQSTCK}{\textit{MQ}$_{+Stack}$\xspace}

\newcommand{\GeneRAG}{\textsc{GeneRAG}\xspace}
\newcommand{\FrostTolerance}{\textsc{FrostTolGene}\xspace}
\newcommand{\OzWheat}{\textsc{OzWheatGene}\xspace}

\newcommand{\BioASQ}{\textsc{BioASQ-QA}\xspace}

\newcommand{\QrQ}{$Qr2Q$\xspace}
\newcommand{\AQ}{$A2Q$\xspace}
\newcommand{\CQ}{$C2Q$\xspace}
\newcommand{\AC}{$A2C$\xspace}
\newcommand{\CAb}{$C2Ab$\xspace}

\newcommand{\Punif}{$\textit{Met}_{unif}$\xspace}
\newcommand{\Prand}{$\textit{Met}_{rand}$\xspace}
\newcommand{\Pdmw}{$\textit{Met}_{dir+hum}$\xspace}
\newcommand{\Plmw}{$\textit{Met}_{log+hum}$\xspace}
\newcommand{\PLdmw}{$\textit{Met}_{dir+llm}$\xspace}
\newcommand{\PLlmw}{$\textit{Met}_{log+llm}$\xspace}

\newcommand{\LLMDir}{\textsc{LLMDirect}\xspace}
\newcommand{\LLMDomain}{\textsc{LLMDomain}\xspace}
\newcommand{\LLMGeneric}{\textsc{LLMGeneric}\xspace}
\newcommand{\LLMAllcheck}{\textsc{LLMAll}\xspace}

\newcommand{\DMW}{\textsc{DMW}\xspace}
\newcommand{\LMW}{\textsc{LMW}\xspace}

\usepackage{tikz}
\usetikzlibrary{shapes.geometric, arrows, calc}
\usetikzlibrary{shapes.geometric, arrows.meta, positioning}
\tikzstyle{process} = [rectangle, rounded corners, minimum width=3cm, minimum height=1cm,text centered, draw=black, fill=blue!20]
\tikzstyle{arrow} = [thick,->,>=stealth]

\title{Improving the Evaluation of Biomedical Question Answering with Few Human Preference Signals}

\author{
  Shakila Mahjabin Tonni$^{1*}$ \quad
  Xiang Dai $^{2}$ \quad
  Sarvnaz Karimi$^{3}$ \\
  \textbf{Xiaomei Li$^{4}$} \quad
  \textbf{Samuel C. Andrew$^{2}$} \quad
  \textbf{Stephen Wan$^{2}$} \\
  $^{1}$Suncorp Bank, Australia \quad
  $^{2}$CSIRO, Australia\\\quad
  $^{3}$Monash University, Australia \quad
  $^{4}$Commonwealth Bank, Australia \\
  \texttt{$^{*}$shakila.tonni@suncorp.com.au}
}

\begin{document}
\maketitle
\begin{abstract}


Retrieval-Augmented Generation (RAG) systems are increasingly used for biomedical Question Answering (QA). However, evaluating their effectiveness remains challenging due to the lack of reliable reference resources and the frequent misalignment between automatic evaluation metrics and human judgments. 
One practical question is how evaluation results should be used to guide development decisions, especially when they correlate weakly with human judgments and may even lead to different conclusions.
We explore how to use a small amount of human preference signals to calibrate direct assessment evaluation metrics, 
%
and show that calibrated metrics consistently outperform baselines that either treat all metrics equally or rely on a single best metric across multiple datasets.

\end{abstract}

\section{Introduction}
\label{sec:intro}

Large Language Models (LLMs)~\citep{biogpt2022, bolton2024biomedlm, labrak2024biomistral, kafkas2025} have shown promise in answering biomedical questions. Interest in Question Answering (QA) applications using Retrieval-Augmented Generation (RAG) techniques has also been on the rise \citep[for example]{gpgpt2024, genegpt2024jin, selfbiorag2024, wang2024biorag}.  As a result, a recent direction has been to examine the use of RAG systems for discovery-oriented scientific questions that require long-form responses.
A key problem, however, is the evaluation of their effectiveness, which remains underexplored.

Consider the question  \emph{``What proportion of Transcriptomic studies on plants found ICE1 Transcription factors to be associated with cold acclimation?''}~as an example.  Acquiring ground-truth answers for comparison is often infeasible due to the high cost of knowledge acquisition. 
In the absence of ground truth references, metric-based approaches to evaluation are a logical option.  However, a naive use of metrics is not straightforward.
When applying existing reference-free direct assessment evaluation metrics to evaluate domain-specific QA, e.g., measuring response  
faithfulness or entailment with relevant resources~\citep{schimanski2024towards}, previous studies have reported inconsistent findings regarding the effectiveness of individual metrics~\citep{roychowdhury-soman-2024-evaluation-rag-metrics,oliva-2025-illusion}. 
Furthermore, we highlight that metrics often lead to inconclusive assessments (\S\ref{sec:metric_vs_dimension}) when applied to evaluate individual components of a RAG-based QA system. This raises the question of how to interpret such evaluation results to guide development decisions (e.g., should different metrics be weighted equally and combined, or should a single best metric be identified?).

Beyond problems with the application of metrics, \citet{bavaresco-2025-llms} showed that domain experts apply stricter domain-specific criteria than automated metrics when evaluating LLM responses.  Additionally, direct assessment itself is challenging for experts compared to pairwise comparison or reference-based assessment~\citep{novikova-2018-rankme}. 
To this end, we work with biologists to collect human preferences for long-form answers from a RAG-based QA system and investigate how to calibrate the weights of direct assessment metrics from collected preferences.
We find that a small amount of human preference signals is highly effective for calibrating weights, making the calibrated metrics substantially better than baselines with uniform or random weighting. 
In contrast, calibration based on LLM preferences, used as a low-cost proxy for human preferences, did not significantly improve alignment with human judgments.

Our contributions are: (1) a preference-guided framework for calibrating weights of direct assessment evaluation metrics using correlation analysis and logistic regression; (2) two human preference datasets \FrostTolerance and \OzWheat, for RAG-based QA in plant and agricultural genomics, experiments on which (and an existing \SciArena benchmark) show that calibrated metrics align better with human judgments than uniform-weight or single-metric baselines; and (3) an exploration of LLM-based pairwise preference collection, as a cheap proxy, using generic and domain-specific rubric-based judges.

\section{Problem Statement}

We consider a 3-stage RAG-based system: (1) query formulation from user question; (2) retrieval and corresponding creation of content surrogates (i.e., summarizing retrieved documents) for inclusion in the prompt for answer generation; and (3) LLM answer generation.  In an applied setting, decisions are needed about which variation is most effective for the proposed use case.  Such decisions relate to the choice of LLM for each component or to algorithmic variations in any of the three above‑mentioned steps.

Building on the RAGAS evaluation framework proposed by~\citet{es2024ragas}, we consider the scenario where ground‑truth answers are unavailable, yet a practical evaluation approach is still needed to inform design decisions for the RAG‑based QA system. 
Here, we first collect a pool of answers for a given question by running different variations of a QA system.
Then, using existing automatic evaluation metrics, we rank the candidate answers and select a few 
candidates for human judgment.
Finally, the collected human judgments and feedback are used to build better evaluation metrics and to inform design decisions for the QA system.

\begin{figure}[tb]
    \centering
    \includegraphics[width=\columnwidth]{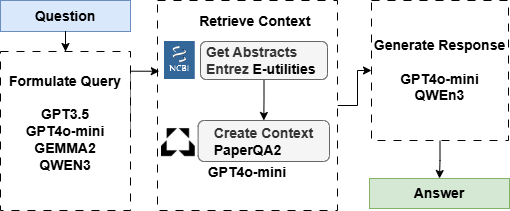}
    \caption{An overview of our RAG-based QA system.}
    \label{fig:qa_process}
\end{figure}

\section{RAG-based QA System}
To enable our study, we first built an in‑house RAG QA system, \GeneRAG, 
since the existing biomedical QA systems are either not publicly available or focused solely on generating short answers.
Figure~\ref{fig:qa_process} shows the pipeline of our system and design options for each stage.

\paragraph{Query Formulation.} 
Our focus user group, consisting of plant genomics researchers, typically relies on NCBI's Entrez system\footnote{\url{https://www.ncbi.nlm.nih.gov/search/}} to search databases (e.g., PubMed) for relevant scientific information. To search via the Entrez API, we first convert a user question to search terms.
To do this, we prompt LLMs with few‑shot examples to: (1) generate a single query term for a given question (this setting is denoted as \SQ); and, (2) create sub-questions from the given question and then generate a query term for each sub-question (denoted as \MQ). 

\paragraph{Context Retrieval.} We retrieve relevant abstracts from \PubMed focusing only on the open-access subset. A brief description of the API is provided in the App. \S\ref{sec:app_entrez}. 
In the \MQ setting, where multiple query terms are generated, two search variants are implemented in \GeneRAG: (1) concatenating all query terms into a single query string using the {\tt OR} operator (\MQCAT); and (2) retrieving abstracts separately for each query term and concatenating them (\MQSTCK). 
After retrieving the abstracts, we follow \PaperQA~\citep{skarlinski2024paperqa2} pipeline, using \textit{GPT4o-mini} to chunk, filter, re‑rank, and summarize them. The final context is a set of summarized relevant excerpts from the abstracts.

\paragraph{Answer Generation.} Given the user's question and the context from the previous step, the system prompts an LLM to generate the final answer.

\paragraph{Models Selection.} In the \SQ setting, we use four LLMs: \textit{GPT3.5-turbo}, \textit{GPT4o-mini}, \textit{GEMMA2-9B}, and \textit{Mistral-Nemo-12B}. In the \MQ setting, we use \textit{GPT4o-mini} and \textit{Mistral-Nemo-12B}. For answer generation, we use \textit{GPT4o-mini} and \textit{QWEN-3-8B}. We evaluated several domain-specific (BioMistral-7B, MedLLAMA-7B, PMC-LLAMA-7B and  BioGPT-Large) and generic LLMs (QWEN2.5-32B and Claude-3-Sonnet) 
on a question subset, selecting models via a quality control inspection.

\section{Datasets}
\label{sec:dataset_and_annotation}
In collaboration with two plant genomics scientists, we compiled two new datasets. 
The scientists selected candidate wheat genes identified through their own genomic annotation research and posed questions regarding the association of these genes with specific traits.
The resulting questions are divided into two sets (examples in App. Table~\ref{tab:qa_sample}): 
\begin{itemize}
    \item \FrostTolerance, a set of 42 questions 
    about the relevance of genes to frost tolerance and cold acclimation, using genes 
    identified through a literature search by the domain experts. 
    \item \OzWheat, a set of 23 questions from genome-wide association analysis that identified genes linked to crop yield and component traits for crop yield for the genes from \textit{Ozwheat} diversity panel~\citep{hyles2024ozwheat}. 
    These questions are related to multiple genes and physical traits of wheat.
\end{itemize}



\subsection{Collecting Human Preferences}
\label{sec:human_annot_setup}
We collected expert preferences over pairs of answers from the \FrostTolerance and \OzWheat datasets through an internally developed survey application (details in App.~\S\ref{sec:app_human_annot_app}). For each question, we generated a pool of 12 answers using different variants (i.e., different combinations of RAG system parameters) using our QA system.

In cases where the answer space is combinatorial, it is common to sample and choose candidate LLM answers for human annotations, given the complexity and time requirement. For example, \citet{feng-2025-sample} selected the top 10 candidates via maximum-discrepancy competition and WildBench \citep{lin2025wildbench} discarded highly similar chats from evaluation. 
Furthermore, \citet{pan2026matters} showed that in the Direct Preference Optimization (DPO) context, the human preference data should cover both good and bad responses. 
Considering the substantial human annotation effort required to judge pairwise answers with large retrieved contexts, we also sampled candidate answers from the answer pool (66 unique combinations of answer pairs per question).
Given our focus on calibrating metric weights for only a specific set of metrics, we used the aggregated evaluation scores as the basis for selecting three candidate answers per question, corresponding to the highest-, moderate-, and lowest-quality answers.

We adopted a tournament-style sampling strategy, 
where the winner from the first random pair moves forward and is paired with the remaining third answer, as we found when comparing annotations on `all' pairs and fewer pairs in tournament-style setup (paired randomly), they both achieve close inter-annotator agreements Cohen's $\kappa$ of 0.68 and 0.64, respectively. 
Critically, the annotators were not aware of the quality ranks of the answers. 
During the annotation process, the annotators could only see the generated contexts and the answers, and did not have access to the formulated query or the retrieved set of abstracts. In addition to selecting their preferred answer, the annotators were instructed to provide a short explanation of the problems they identified in the unpreferred answer (referred to later as the \textit{human reasonings}). 


Of the 65 questions across both datasets, two questions (one from each dataset) were reserved for annotator training and excluded from analysis. For the remaining 63 questions, we collected annotations on answer pairs, resulting in 166 pairwise comparisons. Three domain expert annotators annotated ten questions, and the remaining 53 were single-annotated. For the multi-annotations, we further considered the mode over the multi-annotations and concatenated the reasons. The final data set totalled 130 unique annotations.


\subsection{Preferences by LLM-as-a-Judge}
To compensate for the high cost of human expert annotations, which invariably limits the number of collected human annotations, we developed a rubric-based LLM-as-a-judge baseline \citep{hashemi-2024-llm, lin2025wildbench, zhao2025sciarena} and constructed a synthetic dataset containing LLM preferences over all pairwise answers. Following \citet{lin2025wildbench}, we first created a set of generic rubric questions and a set of domain-specific rubric questions in consultation with domain experts. We then prompted a \textit{GPT-4o-mini} model to give preferences over pairwise answers using the rubric as a guideline. 
We first collected LLM preferences over the human-annotated subset under four settings: \LLMDir---direct LLM preference without rubrics; \LLMGeneric---rubrics containing only generic questions; \LLMDomain---rubrics containing only domain-specific questions; and \LLMAllcheck---rubrics containing both generic and domain-specific questions. The prompts are included in App. ~\S\ref{sec:app_llm_judge}. 
After generating the preferences, we measured human--LLM agreement using Cohen's $\kappa$ and correlation using Pearson's $r$. As shown in Table~\ref{tab:hum_llm_agree}, the \LLMDomain achieved the highest agreement and correlation with human preferences. Therefore, we used the \LLMDomain prompts to collect 4,290 LLM preferences over all answer pairs for each question.



\subsection{Inter-annotator Agreements} 

We computed agreement on the common-answer pairs shared across human annotators. The overall agreement level is moderate with $\kappa$ =0.68. 
The agreements between the annotator pairs are 0.53, 0.66 and 0.72, indicating that the annotators have higher variability in individual alignment with the remaining annotators. This also highlights the challenge of conducting human evaluation in highly specialized domains, since domain experts usually have slightly different background knowledge, which can strongly influence their judgments. 

\begin{table}[tb]
\small
\centering
\resizebox{\columnwidth}{!}{%
\begin{tabular}{@{}lll@{}}
\toprule
Quality       & Metric                                      &  Evaluation    \\ 
Dimension                      &                                              & Dimension\\\midrule

Semantic   Similarity   & ROUGE-1 \citep{lin2004rouge}, & \QrQ,     \\
                        & BERTScore \citep{tianyi2020bertscore}, & \CAb, \AC \\
                        & CosSim. \citep{reimers2019sentence} &    \\\midrule
                        
Relevance               & Context Relevance \citep{es2024ragas} & \CQ \\\cmidrule{2-3}
(LLM-based)             & Answer Relevance  \citep{es2024ragas} & \AQ \\\midrule

Factual consistency & FactCC \citep{kryscinski-2020-evaluating}  & \CAb, \AC \\\cmidrule{2-3}
                    & AlignScore \citep{zha2023alignscore}      & \QrQ, \CQ, \AQ, \\
                    &                  &  \CAb, \AC \\\midrule
Faithfulness        & HHEM-2.1-open \citep{es2024ragas}    & \CAb, \AC \\
 \bottomrule 
\end{tabular}%
}
\caption{Evaluation metrics selected for different evaluation dimensions. \QrQ=query-to-question, \CQ=context-to-question, \AQ=answer-to-question, \CAb=context-to-abstract and \AC=Answer-to-Context.}
\label{tab:metric_list}
\end{table}

\section{Automatic Evaluation vs. Preferences}

Drawing from existing evaluation frameworks in the literature, we first identified a set of quality dimensions and corresponding metrics for the evaluation of RAG-based systems. Beyond assessing only the quality of final answers, following~\citet{es2024ragas}, we used these metrics to assess individual components of the \GeneRAG pipeline. 

\begin{table}[]
\centering
\resizebox{0.7\columnwidth}{!}{%
\begin{tabular}{@{}lcc@{}}
\toprule
Prompt Strategy & Cohen's $\kappa$ & Pearson's $r$           \\ \midrule
\LLMDir         & 0.46          & 0.409\\
\LLMDomain      & \bf 0.55          & \bf 0.530\\
\LLMGeneric     & 0.50          & 0.486\\
\LLMAllcheck    & 0.53          & 0.515\\
\bottomrule
\end{tabular}%
}
\caption{Agreement between human vs. LLM preferences in four LLM-as-a-Judge settings (p-val<.001).}

\label{tab:hum_llm_agree}
\end{table}
\subsection{Metrics and Dimensions}
\label{sec:metric_vs_dimension}

We applied multiple metrics 
across five evaluation dimensions. Specifically, we assessed the quality of the formulated query, the retrieved context, and the final answers to the user questions, which we refer to as in the \QrQ, \AQ and \CQ evaluation, respectively. Additionally, in \AC evaluation, the generated answers were checked against the context, and in \CAb evaluation, the generated contexts by \PaperQA were evaluated against the retrieved abstracts. 
We selected the semantic similarity, relevance, factual consistency and faithfulness quality dimensions and adopted suitable metrics for each dimension. The metric list is presented in Table~\ref{tab:metric_list}. Altogether, we considered 18 evaluation metrics and applied min-max scaling to bring the scores within a range between 0 and 1.  
(Details in App.~\S\ref{sec:app_baseline_metrics}.)

\begin{table}
\centering
\resizebox{\columnwidth}{!}{%
\begin{tabular}{@{}lllccccc@{}}
\toprule
Setting &Query&Ans & \QrQ & \CQ & \CAb & \AQ& \AC\\ 
  & LLM & LLM &Avg. &Avg. &Avg. &Avg. & Avg.\\\midrule
\SQ     & GPT3.5 & GPT4 & \underline{0.606} & 0.154 & 0.465 & 0.793 & 0.408\\
        & GPT3.5  & QWEN & & & &0.813 & \bf 0.497\\\cmidrule{2-8}
        
 & GPT4    & GPT4  & 0.596 & 0.175 & 0.486 & 0.789 & 0.414 \\
 & GPT4    & QWEN & & & & \bf 0.823 & \underline{0.484}\\\cmidrule{2-8}

 & GEMMA   & GPT4  & 0.596 & 0.183 & 0.471 & 0.774 & 0.422 \\
 & GEMMA   & QWEN & & & & 0.804 & 0.478 \\\cmidrule{2-8}
 
 & Mistral & GPT4  & 0.567 & 0.220 & \underline{0.481} & 0.792 & 0.391 \\
 & Mistral & QWEN & & & & 0.811 & 0.473 \\\midrule
 
\MQCAT  & GPT4   & GPT4 & \bf 0.607 & 0.264 & \bf 0.521 & 0.788 & 0.392  \\
        & GPT4   & QWEN & & & & 0.779 & 0.399 \\\cmidrule{2-8}
        
 & Mistral & GPT4 & 0.564 & 0.268  & 0.517  & 0.807 & 0.421  \\
 & Mistral & QWEN & & & & 0.785 & 0.433 \\\midrule
 
\MQSTCK & GPT4   & GPT4 & 0.579 & \bf 0.313 & 0.511 & 0.804 & 0.399 \\
        & GPT4    & QWEN & & & & 0.748 & 0.423  \\\cmidrule{2-8}
        
        & Mistral & GPT4  & 0.578 & \underline{0.312} & 0.511 & \underline{0.812} & 0.399 \\
        & Mistral & QWEN & & & & 0.743 & 0.422  \\
\bottomrule
\end{tabular}%
}
\caption{Metric scores for \FrostTolerance query, context and answer evaluation in three settings: \SQ, \MQCAT and \MQSTCK (Best \textbf{bolded}, and second-best \underline{underlined}).}

\label{tab:frosttolerance_summary_metrics}
\end{table}




From the metric scores, we found that none of the variants \textit{alone} is consistently performing well across different evaluation dimensions, and it is not possible to conclusively decide which LLMs and design options to use in the various stages of the RAG system based on them. 
%
%
For example, summarized results on \FrostTolerance in  Table~\ref{tab:frosttolerance_summary_metrics} shows, the best context was obtained by GPT4 in the \MQSTCK setting for \CQ, but for \CAb, 
\MQCAT setting obtained the highest results. More results and discussion are in the App.~\S\ref{sec:app_metric_results}.
%
This raised our next question: if these metrics cannot reach consensus, how can we use them to guide design decisions for the QA system?

\subsection{Metric vs. Preferences}
\label{sec:metric_vs_human}

We started by analysing the alignment of the human preferences and aggregated metric scores on \FrostTolerance and \OzWheat datasets. We converted 
the scores and the human into binary preference labels. Formally, for each question $q$ and answer pair 
$(a, b)$, a label of 0 is assigned if annotators prefer $a$, and 1 if they prefer $b$. Metric-based preferences are derived analogously:
\begin{equation}
\label{eq:pref_label}
Pref(a, b| \mathcal{S}) =
\begin{cases}
0, & \text{if } \mathcal{S}(a) \geq \mathcal{S}(b), \\
1, & \text{otherwise}.
\end{cases}
\end{equation}
\noindent
where uniform weight $\mathcal{S}(a_i) = \frac{1}{n} \sum_{j=1}^{n} M_j(a_i)$ calculated over $M_j \in \mathcal{M}$ metrics.
Using these labels, we computed the rate of cases when human annotators selected metric-based preferred answers. All metrics were considered equally, equivalent to an \textbf{uniform-weighting}. The resulting agreement percentages and Pearson's correlation coefficients $r$, along with the statistical significance reported in App. Table~\ref{tab:annot_metric_agreement_rates}.
Overall, experts agreed with the metric-based preferences in 61.45\% of the cases (Kappa $\kappa$ = 0.30). 
We also measured a moderate but significant Pearson correlation coefficient, $r$ = 0.300 
(p < 0.01) between human-metric preferences.

Further, between human and individual metrics-based preferences, we found that the agreement rates vary substantially across metrics. 
We found \AC-\textit{AlignScore} has the highest human-metric preference agreement of 65.75\%.  Additionally, \AQ-\textit{AlignScore} achieved 61.64\% agreement, \CAb-\textit{FactCC} 61.45\% and \AC-\textit{Faithfulness} achieved 61.64\% human-metric preference agreements.

In light of the findings, we infer that not all automatic metrics correlate uniformly with human preferences, and it is not possible to have a conclusive decision on the metrics to choose. 
Thus, in the next section, we explore ways to incorporate human preferences into metric reliability calibration.

\section{Preference-Guided Weight Calibration}
\label{sec:met_calib}
Before applying or aggregating evaluation metrics, we want to determine which metrics align better with human preferences. 
The human–metric preference agreement rates computed using uniformly weighted metrics in the previous section allow us to identify which metrics contribute more strongly to alignment with human judgments. These agreements can be considered as a proxy for human preference alignments. 


Based on this, we propose a preference-guided approach for calibrating metric reliability weights using human preferences. Formally, let the agreement rate between human and metric preferences for metric $M_j$ be denoted as the calibration weight $w_j$. Using these weights, the adjusted aggregated score for an answer $a_i$ is computed as $\mathcal{S}_{adj}(a_i) = \frac{1}{n} \sum_{j=1}^{n} w_j \times M_j(a_i)$. The corresponding preference labels, $Pref(a,b \mid \mathcal{S}_{adj})$, are then obtained using Eq.~\ref{eq:pref_label}.
%
%
We consider two approaches to calibrate the metric weights---directly using human–metric preference agreement as weights (\DMW), and learning the weights using a logistic regression model (\LMW). 

\paragraph{Direct Metric Weighting (\DMW)} 
We considered human-metric preference agreements directly as the metric weights. By dividing the dataset into a calibration and validation set and computing human–metric preference agreement scores for each metric on the calibration set, we obtained the calibrated weights for the metrics. We then applied these weights to the validation set for aggregating the metric scores and obtaining the weighted-metric preferences, \Pdmw. Finally, we measured the human-\Pdmw agreement rate and compared it with the agreement rates between human vs. uniform- and random-weighted metric preferences, \Punif and \Prand, respectively. Similarly, we calibrated weights using LLM preferences and obtained metric preferences \PLdmw over the human validation sets and measured the agreement rates against human preferences. 

\paragraph{Learned Metric Weighting (\LMW)} To learn the metric weights from the human preferences, we built separate logistic regression models for each metric. For each metric, we prepared the dataset with pairwise high and low metric scores (implying metric-based preference) for each answer pair, and human preference [0,1] as the label. The model's prediction task was essentially a binary classification task to predict human preference for a pair of high-low metric scores. We then considered the training accuracy of the model as the metric weight. The learned metric weights were then applied to the human validation subsets to obtain \Plmw and validated against the human preferences.

\begin{table}
\centering
\resizebox{0.7\columnwidth}{!}{%
\begin{tabular}{@{}lcc@{}}
\toprule
\multicolumn{3}{c}{\textbf{No Calibration}} \\\midrule
Uniform weighting & \multicolumn{2}{l}{61.44 $\pm$ 5.95} \\
Random weighting & \multicolumn{2}{l}{55.42 $\pm$ 7.21} \\\midrule
\multicolumn{3}{c}{\textbf{W/ Calibration}}\\\midrule
 & \DMW & \LMW\\\midrule
Human pref.& 64.30 $\pm$ 4.34 & 70.00 $\pm$ 4.05\\
LLM  pref.& 53.17 $\pm$ 2.45 & 62.95 $\pm$ 4.10\\
\bottomrule
\end{tabular}%
}
\caption{Agreement rate (\%) and std. dev. (CI=95\%) between human and metric-based preferences, computed via Direct Metric Weighting (\DMW) and Learned Metric Weighting (\LMW), and compared against random and uniform weights on the validation set.}
\label{tab:calib_comparison}
\end{table}





\subsection{Experimental Setup}


We randomly split the human-annotated dataset into a 60\%-40\% calibration–validation split, yielding 78 calibration and 52 validation instances per set, repeating this process 100 times to report the mean and standard deviation of the agreement rates of the human-metric preferences. 
%
%
For experiments involving LLM-generated preferences, we randomly sampled a larger calibration set by taking 80\% of the annotations to derive the calibrated weights for \(k = 100\) iterations in both \DMW and \LMW settings. For \DMW, we averaged the 100 calibration weights to generate a single set of weights. Validation was done on the 100 human-annotated validation folds described above. 
We also considered two baselines: uniform weighting and random weighting, where all metrics are assigned equal weights or random weights, respectively.


\subsection{Results and Discussion}
\label{sec:calib_results}

\paragraph{Calibration with Human Preferences.}
Human preference-based calibration leads to a significant improvement in alignment between human and metric preferences on the validation set, with an increase in the average agreement rate from 61.44\% with uniform metric-weights to 64.30\% with \DMW and 70.00\% with \LMW methods (Table~\ref{tab:calib_comparison}). 
This result is also significantly higher than the \textit{random-weights} baseline with an agreement level of 55.42\%.


We observed a moderate Pearson's correlation coefficient $r = 0.315$ 
for \Pdmw-human and $r = 0.408$ for \Plmw-human preference labels, 
compared to a lower correlation of $r = 0.281$ 
for \Punif-human labels across all $k$ validation samples (for all, p<0.0001). 
%
Furthermore, we found that the improvement in the human-metric agreement rates is also statistically significant. With a one-tailed (Group A > Group B) Wilcoxon’s signed-rank test~\citep{wilcoxon2006}, we measure the agreement rates using \DMW  vs. uniform-weighting over $k$ samples and observed a statistically significant improvement $z = 2.865$ (p<0.01, significant at $\alpha = 0.005$ level)
and which is for \LMW, $z = 0.178$ (p<0.01, significant at $\alpha = 0.05$ level), implying a positive significance with a tighter bound of standard deviations. 


\paragraph{Comparison of Best Metrics and Subsets.} To decide on which evaluation dimensions were most effective in weight calibration, we use \QrQ, \CAb, \CQ+\CAb, \AQ+\AC subsets separately and 
the best metric, \AC-AlignScore (based on App. Table~\ref{tab:annot_metric_agreement_rates}) to calibrate weights. The results are reported in Table~\ref{tab:metric_ablation}. Under the \DMW setting, the single best metric achieved 64.19\%, which was close to the 64.30\% obtained when using all metrics, implying that under direct metric calibration, considering one best metric is sufficient. In contrast, under the \LMW setting, neither the best-performing metric nor any individual subset of metrics achieved a score close to the 70\% obtained using metrics from all dimensions. Furthermore, the \QrQ, \AQ, and \AC metrics appear to contribute the most toward learning the metric weights. 

\paragraph{Calibration with LLM-preferences.} We found that, in the absence of human-annotated data, LLM-generated preferences can also be used to calibrate metric weights. As shown in Table~\ref{tab:calib_comparison}, the \DMW\ setting achieves only 53.17\% agreement with human judgments, which is lower than both uniform and random-weighted baselines. In contrast, the \LMW\ setting improves this agreement to 62.95\%. Although calibration using LLM preferences does not reach the 70\% agreement achieved with human-preference-based calibration, it provides a better alternative to uniform metric aggregation when human annotations are unavailable. 
Across all $k$ samples, we also observe a Pearson correlation of $r = 0.293$ (p < 0.001) between human-\PLdmw\ labels, and $r = 0.396$ (p < 0.001) between human and \PLlmw\ labels, further indicating that LLM-preference in \LMW\ is relatively closer to human-preference-based calibration.

\begin{table}[]
\centering
\resizebox{0.7\columnwidth}{!}{%
\begin{tabular}{@{}llll@{}}
\toprule
   Eval. Dimension  & Uniform & DMW   & LMW    \\ \midrule
All  & 61.44 & 64.30 & 70.00\\\midrule
\AC-AlignScore (best) & \textbf{66.03} & \\
\hline
\hline
\QrQ     & 50.92   & \textbf{52.88} &  \textbf{52.17} \\\hline
\CQ      & \textbf{52.84}   & 52.85 & 51.88  \\
\CAb     & 45.07   & 48.54 & 47.64  \\
\CQ+\CAb & 48.57   & 49.57 & 48.66  \\\hline
\AQ      & 48.19   & 50.80 & 46.58  \\
\AC      & \textbf{61.15}   & \textbf{62.03} & \textbf{58.85}  \\
\AQ+\AC & \textbf{58.54}   & \textbf{60.35} & \textbf{61.55}  \\ 
\bottomrule
\end{tabular}%
}
 \caption{Human-metric preference agreements(\%) for calibration using subsets of metrics compared to `All' metrics. The top 4 best-performing subsets are bolded for each setting.}
\label{tab:metric_ablation}
\end{table}

\paragraph{Evaluation on the \SciArena Benchmark.} To test the generalizability of the calibration techniques on an out-of-domain dataset, we conducted experiments on \SciArena \citep{zhao2025sciarena}, which is a benchmarking dataset 
with annotations collected from the researcher community over pairwise LLM-generated responses on scientific literature tasks. We considered \textit{`Biomedical Engineering'}, \textit{`Medicine'}, \textit{`Neuroscience'}, \textit{`Pharmacy'} and \textit{`Public Health'} subsets, as the domains were close to our focus. Details of the data size of each subset are reported in the App.~\S\ref{sec:app_sciarena}, and an example QA pair is in Table~\ref{tab:sciarena_exmpl}. We applied both \DMW and \LMW techniques on each subset separately and combined. Results are in Table~\ref{tab:sciarena_agreements}. We observe that in all cases human-metric agreement rates improved from the uniform weighting, and the best agreement rate of 60.51\%, on the combined set, was achieved in the \LMW setting. Note, in the SciArena dataset, only answers and context (citations) are available, and we used only 9 metrics covering \CQ,\AQ and \AC.  
As shown in \S\ref{sec:calib_results}, the \LMW setting performed better with a larger set of metrics, particularly with \QrQ, \AQ, and \AC, which might be a possible reason for the improvement in agreement from \DMW to \LMW.

\begin{table}
\centering
\resizebox{0.8\columnwidth}{!}{%
\begin{tabular}{lccc}
\toprule
              & \multicolumn{1}{l}{Unif. Weighting} & DMW    & LMW    \\\midrule
Combined &  53.47                                & 59.82 & \bf 60.51 \\\midrule
Biomed. &   52.14                               & \bf 60.17 & 59.82 \\
Medicine&   53.91                                & \bf 58.77 & 58.75 \\
Neuroscience& 54.00                                & 61.09 & \bf 62.18 \\
Pharmacy&   52.42                                & 60.05 & \bf 60.57 \\
Public Health&  51.77                                & 59.00 & \bf 61.23\\\bottomrule
\end{tabular}%
}
\caption{\SciArena human-metric preference agreements(\%).}

\label{tab:sciarena_agreements}
\end{table}

\paragraph{Quality Dimensions from Expert Reasons.} 
To understand which quality aspects the experts prioritize, we prompted a \textit{GPT4o-mini} model with a one-shot example to group the human rationals and assign interpretable labels (referred to as \textit{criteria}). In case multiple reasons were present in one rationale, we split and consider them separately. Using another prompt by adding these dimensions in-prompt, we asked the LLM to assign a label to each of the human reasons, which were further corrected through manual checking. 
The prompt and resulting criteria are in App.~\S\ref{sec:app_quality_dim_analysis}. 
We found that the main factors influencing human judgments are answer and context relevance, lack of detail in the answer and incorrect or irrelevant context (App. Table~\ref{tab:app_gpt4_dimensions}). This validates our contextual appraisal of quality based on a set of component-wise metrics for RAG systems. 

We also observed 35 cases where annotators noted that the answers were very similar, resembling a `tie' between two answers and that both answers are good/bad~\citep{chatbotarenav235chiang24b}. 
Two such examples are in App. Table~\ref{tab:app_annot_exmpl1}.

In addition, human experts also considered answer clarity, fluency, organization, and structure. 
This finding resonates with prior work that incorporates stylistic qualities~\citep{dsouza-2025-yescieval} 
and answer usefulness~\citep{malaviya2024expertqa} in LLM-based evaluation. 

We further analyzed cases where metric-based preferences differ from human judgments (App. Table~\ref{tab:app_dim_mismatched}), highlighting the metrics’ inability to cover certain evaluation criteria. For example, when only \CAb metrics are used to derive metric preferences, they disagree with 30 human preferences in cases where humans based their judgments on `Relevance of Answer'. 
The lowest amount of mismatching was obtained when all metrics were considered to generate preferences.

Finally, we observed that the qualitative aspects of the answers can be a matter of subjective judgments, varying across expertise levels. App. Table~\ref{tab:app_annot_exmpl2} gives an example of such an annotation, where both annotators considered different rationales for the same preference. Therefore, effectively selecting an evaluation strategy for RAG systems requires incorporating metrics from all components of the RAG pipeline and human-in-the-loop evaluation strategies.

\par{Practicality and Overall Recommendation.} In our work, we aim to improve the alignment between automatic evaluation metrics and human judgments. In other words, we aim to identify a better and more cost-effective proxy for human judgments to facilitate downstream RAG system development. Our main goal is to show that, even with only a small number of human annotations, they can be used to effectively calibrate the metric weights. The calibrated metric reaches a performance level close to human–human agreement, suggesting that this is a practical upper bound given the variability in human judgments. We also observe that, although, \LMW is better-performing in all cases, \DMW provides only modest improvement and, as we reported in the Table \ref{tab:metric_ablation}, performs worse than the single best metric. This suggests that metric weight calibration is not always needed. Therefore, we recommend comparing the calibrated metric with the single best metric and using whichever achieves higher agreement with human preferences.

\section{Related Work}
\label{sec:app_related_work}

We first explored the limitations of different QA datasets and language models in the biomedical domain. Then, we discussed challenges with the existing automated metrics for RAG-based QA systems and how human preferences are integrated to improve LLM-based evaluations.

\paragraph{Biomedical QA Datasets.} Several general and specialized datasets support biomedical QA. MedQA~\citep{jin2020medqa} and MedRedQA~\citep{nguyen2023medredqa} targeted general medical knowledge, whereas GeneTuring \citep{shang2025geneturing} introduced gene-centric questions, PubMedQA~\citep{jin2019pubmedqa} and \BioASQ~\citep{krithara2023bioasqqa} focused on literature-grounded questions.  
More recently, ExpertQA \citep{malaviya2024expertqa} introduced expert-curated questions across multiple domains.  
However, these datasets are not suitable for evaluating long-form, open-ended responses. Our study introduced qualitative aspects-- 
such as relevance and alignment with the question and supporting context--through calibrated metrics by human preferences.

\paragraph{LLMs for Answering Domain-specific Questions.} LLMs have been pretrained or fine-tuned on biomedical corpora to support domain-specific QA \citep{bolton2024biomedlm, biogpt2022, pmcllama2024, labrak2024biomistral, yasunaga2022linkbert}. 
Some models focused on gene-related tasks, such as Geneverse~\citep{liu2024geneverse} 
and GP-GPT~\citep{gpgpt2024}. 
In parallel, RAG has been explored for biomedical QA using literature retrieval~\citep{skarlinski2024paperqa2, patel2026meddiscover}, retrieval-based gap-analysis and knowledge assertion~\citep{medtrust2025rag} or combining databases with external tools~\citep{wang2024biorag}. However, such systems rely on restricted resources, limiting public availability and reproducibility.

\paragraph{Evaluation of RAG QA systems.} Evaluating 
RAG-based QA models remain challenging due to the lack of reliable reference answers. RAGAS \citep{es2024ragas} proposed a reference-free evaluation framework judging faithfulness, relevance, and groundedness. \citet{adlakha-etal-2024-evaluating} evaluated correctness and faithfulness in instruction-following QA and \citet{zha2023alignscore} evaluated factual consistency via AlignScore. Benchmarks such as LitQA2 \citep{skarlinski2024paperqa2} and LabBench \citep{laurent2024lab} assessed QA accuracy and biological reasoning skills, while ExpertQA \citep{malaviya2024expertqa} evaluated qualitative dimensions including usefulness and cite-worthiness. 
However, guidance on metric selection for domain-specific QA remains limited.

\paragraph{Human-preferences in LLM Evaluators.} LLMs were used as automated evaluators to judge AI-generated responses \citep{li2025reliability, hijazi2024using, dsouza-2025-yescieval}. 
However, there are significant gaps between coverage of human-centred criteria by LLMs \citep{miller2025evaluating} or inconsistent correlations with human judgment \citep{oliva-2025-illusion}.
Consequently, several approaches now integrate human feedback into LLM-evaluators for automated evaluation of reasoning chains \citep{jacovi-2024-chain} and to align automated evaluation with human preferences \citep{yue2025qaeval, polo2025bridginggap, ye2025llmprefalignment}.



\section{Conclusions}

Evaluating the effectiveness of RAG-based QA systems for generating long-form answers to biomedical questions remains a challenging task, regardless of the availability of several automatic evaluation metrics. 
Our analysis revealed substantial variability across metrics when evaluating different components of a RAG system. Such variability makes it difficult to interpret results, select appropriate metrics, or identify a single reliable metric for evaluation. Furthermore, we found that no individual metric consistently aligns with expert judgments.

To address this issue, we considered the human–metric preference agreement rates as indicators of each metric’s contribution toward human alignment and proposed calibrating metric weights before applying or aggregating them. 
Our results showed that, without calibration, relying on the single best-performing metric can achieve performance comparable to direct metric weighting (\DMW). However, the strongest alignment with human preferences was achieved when metrics across all evaluation dimensions---query, context, and answer---were considered and the weights were learned using the \LMW\ approach.

We further conducted similar weight calibration using LLM preferences collected through a rubric-based LLM-as-a-judge baseline. The results showed that only the learning-based \LMW\ approach improved human alignment of the metrics over uniform weighting. However, the performance remained well below the human-preference-based calibration. 

Additionally, to conduct the weight calibrations, we create two expert-curated plant genomics QA datasets that address the scarcity of non-human genetics QA benchmarks, and extended the analysis to the \SciArena benchmark dataset.

Future work includes extending this analysis to other domains or task types to test the generality of preference-guided metric calibration and adapting other learning techniques for improved metric weight calibration. In addition, integrating preference-guided metrics into training objectives for RAG-based QA models—through reinforcement learning or self-refinement—could further improve alignment with expert human judgments.



\section{Limitations}
%
As a proof of concept, we propose using a small amount of human preference signals to calibrate existing direct assessment metrics. We select representative metrics that aim to cover different quality dimensions. We do not claim that our choice of metrics is exhaustive. %
Moreover, given the complexity of discovery-oriented scientific questions and long-form answers, these expert annotations are inherently subjective and challenging to scale consistently, which constitutes a key limitation.

\section{Ethics Statement}

This work received ethics approval through CSIRO’s ethics approval process for the collection of human annotations. 
Three domain scientists served as annotators, two of whom are co-authors of this paper. The scientists were recruited via email invitations from within CSIRO and were involved in genomic research. 
Participants were provided with a document describing the project, including the annotation task, data collection and anonymisation procedures, and ethics approval details. Participation was voluntary, and participants were asked to provide informed consent via email before taking part in the study.

\bibliography{anthology-1,anthology-2,custom}
\bibliographystyle{acl_natbib}

\clearpage
\appendix

\section*{Appendix}
\label{sec:appendix}

\section{Entrez E-utils Framework}
\label{sec:app_entrez}
The National Center for Biotechnology Information (NCBI) is a primary repository for genomic data, integrating diverse datasets for genomes, genes, proteins, and more, with analytical and retrieval tools through its website \citep{entrez2012ostell}. The Entrez Programming Utilities (E-utilities) \citep{entrez2012ostell}, a text-based search system, allows easy navigation across biological databases. Entrez offers a stable interface for querying and retrieving data from 38 NCBI databases, including gene records and biomedical literature \citep{Tatusova2010eutils}. We make two API calls on the PMC open access subset of the articles: \textit{E-Search}, to find the \PubMed IDs (PMIDs) of the most relevant articles for a given query, and \textit{E-Fetch}, to extract the abstracts associated with those PMIDs. 
\begin{table}[ht!]
    \small
    \centering
    \resizebox{\columnwidth}{!}{
    \begin{tabular}[width=0.6\columnwidth]{ll}
    \toprule
    API/Method & Parameters\\\midrule
    NCBI API & DB: \PubMed, Sort: relevance \\
    (Entrez)&Min-year: 1990, Max-year: 2024 \\
    &Max-articles:30, Filter: PMC open access\\
    \midrule
    
    \PaperQA & Top evidences: 5, Evidence summary length: 50\\
         &Max answer source: 3, Answer length: 100\\
         &LLM: GPT4o-2024-08-06\\
    \bottomrule
    \end{tabular}
    }
    \caption{Parameter settings for NCBI API and \PaperQA}
    \label{tab:app_parameters}
\end{table}

\begin{table}[ht!]
\small

    \begin{tabular}[width=0.3\columnwidth]{p{0.9\columnwidth}}
    \cellcolor{black}\textbf{\textcolor{white}{Prompt: Query Reformulation---\SQ setting}}
    \\\midrule
    Question: \{question\}
    \\
    \\
    From the above question, your task is to create a search phrase to find relevant articles. The search phrase should contain only one or a few words. Do not include any word that is not present in the question. Do not generate numbers and punctuation. Convert any initials to their full form.
    \\\\
    Below are some examples: \{samples\}
    \\
    However, below query term(s) didn't return any article:\\
    \{query-error\}
    \\
    Your response should only contain the search phrase.\\
    Search phrase:\\
    \bottomrule
    \end{tabular}
    \caption{Few-shot LLM prompt to formulate single query terms (\SQ) to search articles from the NCBI API.}
    \label{tab:app_que_gen_prompt}
\end{table}

\begin{table}[ht!]
\small

    \begin{tabular}[width=0.3\columnwidth]{p{0.9\columnwidth}}
    \cellcolor{black}\textbf{\textcolor{white}{Prompt: Query Reformulation---\MQ setting}}
    \\\midrule
    Your task is to create multiple search phrases to find relevant articles that may assist in answering a question.
    \\
    \\
    1. If the question is simple and asking about one single topic, keep the question as it is. 
    \\
    2. If the question is asking a relation between multiple topics, create sub-questions that will be able to help get the final answer.
    \\
    3. Create the most important search phrases for the questions to be able to search relevant articles.
    \\
    \\
    The search phrase should contain only one or a few words. 
    Do not include any search term that is not present in the question. 
    Do not generate numbers and punctuations, unless present in the question. 
    Try to convert any initials to their full form.\\
    Below are some samples:
    \\
    \\
    Question: \{sample-q\}\\
    Reasoning:  \{sample-sub-questions\}\\
    Search phrases: [\{sample-query-terms\}]
   \\
   \ldots
   \\
    However, the query term(s) didn't return any article:\\
    \{query-error\}
    \\
    Now create {n} search phrases following the above method for the following question.
    Your response should contain the sub-question, if created, and the search phrases.
    \\
    \\
    Question:\{question\}\\
    Your response:\\
    \bottomrule
    \end{tabular}
    \caption{Few-shot LLM prompt to formulate multiple query terms (\MQ) to search articles from the NCBI API.}
    \label{tab:app_multi_query_gen_prompt}
\end{table}

\begin{table}[ht!]
\small

    \begin{tabular}[width=0.3\columnwidth]{p{0.9\columnwidth}}
    \cellcolor{black}\textbf{\textcolor{white}{Prompt: Answer Generation}}
    \\\midrule
    Question: \{question\}
    \\
    \\
    For the above question, we retrieve the following context:
    \\
    Context: \{context\}
    \\\\       
    Try to answer the question with the help of the given context in a maximum \{length\} words with citation in the format (citation key - pmid).
    \\
    \\
    If the context provides insufficient information, try to reply with as much helpful information as you can give.
    \\
    Your response should only contain the answer to the question.
    \\
    Answer:\\
    \bottomrule
    \end{tabular}
    \caption{LLM prompt for generating answers for the given question based on the retrieved context.}
    \label{tab:app_answer_gen_prompt}
\end{table}
\begin{table*}[t!]
\small
    \centering
    \resizebox{\textwidth}{!}{
    \begin{tabular}{>{\raggedright}p{0.3\textwidth}p{0.7\textwidth}}
    \toprule
    Question & Answer (GPT4)\\
    \midrule
    
    \textbf{\FrostTolerance}---What proportion of Transcriptomic studies on plants found ICE1 Transcription factors to be associated with cold acclimation? \textbf{[Contains both gene name and trait]} & The provided context indicates that ICE1 transcription factors are associated with cold acclimation in multiple plant studies. Specifically, Wu (2021) discusses the regulation of cold acclimation pathways by ICE1 and ICE2 in plants (Wu 2021 - 34639192). Additionally, Xu (2015) identifies ICE gene homologs in grapevines that play a crucial role in cold acclimation, highlighting the involvement of VaICE1 as a transcription factor responding to cold stress (Xu 2015 - 25019620). However, the exact proportion of studies reporting this association is not specified in the context. Thus, while ICE1 transcription factors are recognized for their role in cold acclimation, quantifying the proportion of studies is not possible based on the given information.
\\
    \midrule
    \textbf{\OzWheat}---Candidate gene TraesCS4D03G0067100 on genome GCF$\_$018294505.1 has the annotation of being a GAI-like protein 1 gene. Are there any known associations or links between this family of genes and the trait height in wheat research? Please also describe the evidence for the association. \textbf{[Contains both gene name and connection to pre-existing trait]}& 
    Yes, there are known associations between the GAI-like protein family and height traits in wheat. The Reduced height-1 (Rht-B1 and Rht-D1) loci, which are orthologues of the Arabidopsis GAI gene, encode proteins that function as nuclear transcription factors and are implicated in the regulation of plant height by modulating the response to gibberellin, a growth hormone (Peng 1999 - 10421366). The introduction of mutant dwarfing alleles at these loci has led to the development of shorter wheat varieties, significantly contributing to the green revolution's impact on wheat yields (Peng 1999 - 10421366). Additionally, studies of mutant GAI alleles in other crops, such as rice, demonstrate that they also result in reduced gibberellin responses and dwarfism, suggesting that manipulating GAI-like proteins could be a strategy to enhance yield across various species (Peng 1999 - 10421366).    
    \\

    \bottomrule
    \end{tabular}
    }
    \caption{Example gene-trait association questions from the \FrostTolerance and \OzWheat datasets. Answers are generated by GPT4.}
    \label{tab:qa_sample}
\end{table*}

\section{\GeneRAG Settings}
\label{sec:app_generag_setup}
\GeneRAG retrieves the top 30 articles based on their \textit{relevance} to the query terms and provides them to \PaperQA \citep{skarlinski2024paperqa2}, which uses a ranking technique for creating relevant abstract summaries (to use where LLM input context windows are limited). For \MQCAT, we generated 3 search terms and concatenated them into a single string for article search. In the case of \MQSTCK, we formulated 10 search terms per question and found 10 abstracts for each term. Then we stacked them in a single list of abstracts to create the context using \PaperQA. In the context produced by \PaperQA, we have top-3 evidence summaries in \SQ and \MQCAT settings, and top-5 evidence summaries in \MQSTCK setting. 
We are using \PaperQA specifically to find answer-related passages for the answer generation stage. \PaperQA breaks the retrieved abstracts into smaller chunks and ranks them by their perplexity with respect to the question. Then, the top 5 abstract chunks are selected to generate evidence summaries. Finally, the summaries are further filtered out by re-ranking them by an LLM-prompt. The top 3 of the evidence summaries by this re-ranking were then returned as the final context to be provided to the answer generation stage. The parameter settings for both Entrez system and \PaperQA are outlined in Table~\ref{tab:app_parameters}. %

We also provide the used LLM prompts for query reformulation in \SQ and \MQ settings in Table~\ref{tab:app_que_gen_prompt} \& \ref{tab:app_multi_query_gen_prompt} and answer generation in Table~\ref{tab:app_answer_gen_prompt}. We provide some sample questions and answers for the explored datasets in Table~\ref{tab:qa_sample}.

\begin{table}[ht!]
\centering
\resizebox{\columnwidth}{!}{%
\begin{tabular}{@{}lll@{}}
\toprule
\multicolumn{3}{c}{\textbf{Query \& Context (Prompt Setting--Query LLM)}}\\\midrule
Eval. Dim.             & \FrostTolerance  & \OzWheat      \\ \midrule
\QrQ              & \MQSTCK-GPT4             & \SQ-GEMMA         \\
\CQ               & \MQSTCK-Mistral         & \MQSTCK-GPT4      \\
\CAb              & \SQ-GEMMA   & \MQCAT-Mistral          \\\midrule
\textbf{Overall}   & \SQ-GPT3.5   & SQ-GPT3.5                 \\
\midrule
\multicolumn{3}{c}{\textbf{ Answer ((Prompt Setting--Query LLM--Answer LLM)}}\\\midrule
\AQ               & \SQ-GPT3.5-GPT4  & \MQCAT-Mistral-QWEN     \\
\AC               & \MQCAT-Mistral-GPT4 & \MQCAT-GPT4-GPT4  \\\midrule

\textbf{Overlall} & \MQCAT-Mistral-GPT4  & MQCAT-GPT4-GPT4 \\ \bottomrule
\end{tabular}%
}
\caption{Best-performing LLMs for all datasets, based on metric scores computed for multiple evaluation dimensions.}
\label{tab:metric_summary}
\end{table}


\begin{table*}
\centering
\resizebox{\textwidth}{!}{%
\begin{tabular}{@{}llcccc>{\columncolor[HTML]{C8BBFC}}ccc>{\columncolor[HTML]{D6D3D2}}cccccc>{\columncolor[HTML]{D6D3D2}}c>{\columncolor[HTML]{C8BBFC}}c>{\columncolor[HTML]{FC7649}}c@{}}
\toprule
Prompt & Query & \multicolumn{5}{c|}{Query to Ques. (\QrQ) } & \multicolumn{3}{c|}{Context to   Ques. (\CQ) }& \multicolumn{6}{c|}{Context to   Abstracts (\CAb)}& & \\ 
\cmidrule{3-18}
Setting   & LLM & ROUGE-1 & BERTSc. & CosSim. & AlignSc. & Query Avg.  & AlignSc. & Relev. & Avg. & ROUGE-1 & BERTSc. & AlignSc. & FactCC & Faith. & Avg. & Context Avg.& Overall Avg.  \\ 
\midrule
\SQ     & GPT3.5  & 0.072 & 0.877 & 0.817 & 0.659 & \underline{0.606} & 0.202 & 0.106 & 0.154 & \textbf{0.070} & 0.772 & 0.461 & 0.879 & 0.142 & 0.465 &0.376 &0.460 \\
       & GPT4    & 0.062 & 0.877 & 0.782 & 0.663 & 0.596 & 0.227 & 0.123 & 0.175 & 0.063 & 0.788 & 0.487 & 0.880 & 0.210 & 0.486 &0.397 & 0.469\\
       & GEMMA   & 0.068 & 0.869 & 0.824 & 0.623 & 0.596 & 0.217 & 0.149 & 0.183 & 0.065 & 0.781 & 0.460 & 0.883 & 0.165 & 0.471 &0.389& 0.464\\
       & Mistral & \textbf{0.089} & 0.870 & 0.758 & 0.549 & 0.567 & 0.229 & 0.210 & 0.220 & 0.047 & 0.806 & 0.470 & 0.839 & 0.241 & \underline{0.481} &0.406 & 0.464 \\\midrule
       
\MQCAT  & GPT4    & 0.029 & \textbf{0.884} & \textbf{0.843} & \textbf{0.670} & \textbf{0.607} & 0.228 & 0.300 & 0.264 & 0.043 & 0.826 & 0.455 & \textbf{0.935} & \textbf{0.347} & \textbf{0.521} &\underline{0.448} & \textbf{0.505} \\
       & Mistral & 0.041 & 0.870 & 0.781 & 0.564 & 0.564 & 0.206 & \textbf{0.329} & 0.268 & 0.048 & \textbf{0.848} & 0.460 & 0.906 & 0.324 & 0.517 & 0.446& 0.489 \\\midrule
       
\MQSTCK & GPT4    & 0.047 & 0.855 & 0.777 & 0.636 & 0.579 & \textbf{0.431} & 0.195 & \textbf{0.313} & 0.068 & 0.811 & \textbf{0.541} & 0.930 & 0.204 & 0.511 & \textbf{0.454}& 0.499 \\
       & Mistral & 0.046 & 0.855 & 0.777 & 0.635 & 0.578 & 0.430 & 0.194 & \underline{0.312} & 0.067 & 0.811 & 0.540 & 0.931 & 0.207 & 0.511 &\textbf{0.454} & \underline{0.500} \\ 
\bottomrule
\end{tabular}%
}
\caption{Metric scores for \FrostTolerance query and context evaluation. Queries are formulated by LLMs in three settings, and contexts are produced by \PaperQA. Best scores are \textbf{bolded}, and second-best scores are \underline{underlined}.}
\label{tab:frosttolerance_metrics}
\end{table*}

\begin{table*}
\centering
\resizebox{0.8\textwidth}{!}{%
\begin{tabular}{@{}lllcc>{\columncolor[HTML]{D6D3D2}}cccccc>{\columncolor[HTML]{D6D3D2}}c>{\columncolor[HTML]{C8BBFC}}c@{}}
\toprule
Prompt &Query  & Answer & \multicolumn{3}{c|}{Ans. to Ques. (\AQ)} & \multicolumn{6}{c|}{Ans. to Context (\AC)}& \cellcolor[HTML]{FFFFFF}\\ 
\cmidrule{4-13}

 Setting & LLM & LLM & AlignSc. & Relev. &Avg. & AlignSc. & ROUGE-1 & BERTSc. & FactCC & Faith.  & Avg. & Avg. \\\midrule
\SQ     & GPT3.5 & GPT4 & 0.710 & 0.875 & 0.793 & 0.676 & 0.013 & 0.818 & 0.345 & 0.187 & 0.408 & 0.518 \\
 & GPT4    & GPT4  & 0.719 & 0.859 & 0.789 & 0.695 & 0.018 & 0.822 & 0.254 & 0.283 & 0.414 & 0.521 \\
 & GEMMA   & GPT4  & 0.715 & 0.832 & 0.774 & 0.691 & 0.016 & 0.821 & 0.332 & 0.249 & 0.422 & 0.522 \\
 & Mistral & GPT4  & 0.743 & 0.840 & 0.792 & 0.690 & 0.022 & 0.831 & 0.208 & 0.206 & 0.391 & 0.506 \\\cmidrule{2-13}
 & GPT3.5  & QWEN & 0.758 & 0.867 & 0.813 & \textbf{0.754} & 0.012 & 0.805 & \textbf{0.736} & 0.179 & \textbf{0.497} & \textbf{0.587} \\
 & GPT4    & QWEN & 0.753 & \textbf{0.893} & \textbf{0.823} & 0.730 & 0.017 & 0.813 & 0.684 & 0.175 & \underline{0.484} & \underline{0.581} \\
 & GEMMA   & QWEN & 0.760 & 0.847 & 0.804 & 0.710 & 0.013 & 0.814 & 0.629 & 0.226 & 0.478 & 0.571 \\
 & Mistral & QWEN & 0.768 & 0.854 & 0.811 & 0.696 & 0.029 & 0.824 & 0.572 & 0.244 & 0.473 & 0.570 \\\midrule
\MQCAT  & GPT4   & GPT4 & 0.777 & 0.798 & 0.788 & 0.718 & 0.042 & \textbf{0.862} & 0.091 & 0.249 & 0.392 & 0.505 \\
 & Mistral & GPT4  & 0.756 & 0.858 & 0.807 & 0.688 & \textbf{0.043} & 0.858 & 0.265 & 0.252 & 0.421 & 0.531 \\\cmidrule{2-13}
 & GPT4    & QWEN & 0.716 & 0.841 & 0.779 & 0.659 & 0.042 & 0.854 & 0.060 & 0.379 & 0.399 & 0.507 \\
 & Mistral & QWEN & 0.717 & 0.852 & 0.785 & 0.653 & 0.042 & 0.851 & 0.311 & 0.307 & 0.433 & 0.533 \\\midrule
\MQSTCK & GPT4   & GPT4 & \textbf{0.792} & 0.816 & 0.804 & 0.746 & 0.038 & 0.848 & 0.088 & 0.273 & 0.399 & 0.514 \\
 & Mistral & GPT4  & \textbf{0.792} & 0.831 & \underline{0.812} & 0.746 & 0.038 & 0.848 & 0.088 & 0.273 & 0.399 & 0.517 \\\cmidrule{2-13}
 & GPT4    & QWEN & 0.675 & 0.821 & 0.748 & 0.719 & 0.038 & 0.842 & 0.110 & \textbf{0.403} & 0.423 & 0.515 \\
 & Mistral & QWEN & 0.675 & 0.810 & 0.743 & 0.719 & 0.038 & 0.842 & 0.108 & 0.402 & 0.422 & 0.513 \\
\bottomrule
\end{tabular}%
}
\caption{\FrostTolerance metric scores evaluating the answer quality. Answers are generated by GPT4 and QWEN.}
\label{tab:frosttolerance_ans_metrics}
\end{table*}

\section{Metric Details}
\label{sec:app_baseline_metrics}

The evaluation metrics used in this work are as follows:
\paragraph{Semantic similarity} measures how closely a text aligns with a reference text in terms of meaning. We chose 
ROUGE-1 \citep{lin2004rouge} 
(F1-score), cosine similarity based on SentenceTransformer~\citep{reimers2019sentence} embeddings (CosSim), and BERTScore~\citep{tianyi2020bertscore}, 
and applied them to \QrQ, \CAb and \AC evaluation.  
    
\paragraph{Relevance} is an LLM-based metric adopted from the RAGAS framework~\citep{es2024ragas} and measures \textit{Context Relevance} and \textit{Answer Relevance}  
to the questions for \CQ and \AQ evaluations. 
\textit{Answer Relevance} is measured by prompting an LLM 
to create several possible questions that could have led to the given answer. Finally, averagring the cosine similarities between these generated questions and the original question 
to get the score. 
\textit{Context Relevance} is measured by asking an LLM to extract sentences from context that are essential for answering the question and computing the ratio of relevant sentences to total context sentences.
    
\paragraph{Factual consistency}  evaluates whether the claims in the generated answer can be supported by the source material. 
We use two metrics: 
(1) FactCC~\citep{kryscinski-2020-evaluating}, which checks the factual correctness; and (2) AlignScore~\citep{zha2023alignscore} that measures NLI-based textual entailments between two sets of texts. Here, we applied the pretrained BERT-based FactCC model\footnote{\url{https://huggingface.co/manueldeprada/FactCC}} and RoBERTa-based \textit{AlignScore-base} model\footnote{\url{https://huggingface.co/yzha/AlignScore}}. 
%
We applied AlignScore in testing components against the user questions (\QrQ, \CQ and \AQ). For \CAb and \AC evaluation, we applied both FactCC and AlignScore. 


\paragraph{Faithfulness} refers to the extent to which the LLM's output remains grounded in the evidence. 
For the \CAb and \AC evaluations, 
the information in the context and the answer should be supported by the biomedical evidence given in the reference abstracts and contexts, respectively, without adding irrelevant or hallucinated claims. 
%
We used the pretrained hallucination detection model HHEM-2.1-Open model from Vectara\footnote{\url{https://shorturl.at/naxUx}}.

\begin{table*}
\centering
\resizebox{\textwidth}{!}{%
\begin{tabular}{@{}llcccc>{\columncolor[HTML]{C8BBFC}}ccc>{\columncolor[HTML]{D6D3D2}}cccccc>{\columncolor[HTML]{D6D3D2}}c>{\columncolor[HTML]{C8BBFC}}c>{\columncolor[HTML]{FC7649}}c@{}}
\toprule
Prompt & Query & \multicolumn{5}{c|}{Query to Ques. (\QrQ) } & \multicolumn{3}{c|}{Context to   Ques. (\CQ) }& \multicolumn{6}{c|}{Context to   Abstracts (\CAb)}& \cellcolor{white} & \cellcolor{white} \\ 
\cmidrule{3-18}
Setting   & LLM & ROUGE-1 & BERTSc. & CosSim. & AlignSc. & Query Avg.  & AlignSc. & Relev. & Avg. & ROUGE-1 & BERTSc. & AlignSc. & FactCC & Faith. & Avg. & Context Avg.& Overall Avg.  \\ 
\midrule
\SQ     & GPT3.5  & \textbf{0.049} & 0.853 & 0.761 & 0.544 & 0.552 & 0.273 & 0.028 & 0.151 & 0.075 & 0.753 & 0.578 & 0.989 & 0.054 &\underline{0.650} &0.393 & \textbf{0.523} \\
       & GPT4    & 0.046 & 0.839 & 0.696 & 0.483 & 0.516 & 0.280  & 0.085 & 0.183 & 0.032 & 0.760  & \textbf{0.584} & 0.912 & 0.083 & 0.634 & 0.391&0.509 \\
       & GEMMA   & 0.046 & 0.839 & 0.696 & 0.483 & 0.516 & 0.280  & 0.112 & 0.196 & 0.032 & 0.760  & \textbf{0.584} & \textbf{0.995} & 0.083 & \textbf{0.651} &\underline{0.407} & \underline{0.519} \\
       & Mistral & 0.046 & 0.851 & 0.703 & 0.543 & 0.536 & 0.289 & 0.070  & 0.180 & \textbf{0.093} & 0.765 & 0.572 & 0.898 & 0.096 & 0.625& 0.398& 0.511 \\
\midrule
\MQCAT  & GPT4    & 0.036 & 0.858 & \textbf{0.806} & 0.605 & \underline{0.576} & 0.424 & 0.372 & 0.398 & 0.045 & 0.828 & 0.494 & 0.304 & \textbf{0.317} & 0.398 &0.398 & 0.463 \\
       & Mistral & 0.027 & 0.852 & 0.732 & 0.575 & 0.547 & 0.414 & \textbf{0.402} & \underline{0.408} & 0.084 & \textbf{0.830}  & 0.514 & 0.419 & 0.226 & 0.415 & \textbf{0.413}& 0.461 \\\midrule
\MQSTCK & GPT4    & 0.040  & \textbf{0.861} & 0.801 & \textbf{0.657} & \textbf{0.590} & 0.488 & 0.270  & 0.379 & 0.043 & 0.813 & 0.493 & 0.205 & 0.204 & 0.352 & 0.359& 0.443 \\
       & Mistral & 0.036 & 0.840  & 0.716 & 0.647& 0.560 & \textbf{0.560}  & 0.296 & \textbf{0.428} & 0.042 & 0.810  & 0.528 & 0.254 & 0.163 & 0.359 &0.379& 0.445 \\
\bottomrule
\end{tabular}%
}
\caption{Metric scores for \OzWheat query and context evaluation. Queries are formulated by multiple LLM settings and contexts are produced by PaperQA (GPT4).}
\label{tab:ozwheat_metrics}
\end{table*}

\begin{table*}
\centering
\resizebox{0.8\textwidth}{!}{%
\begin{tabular}{@{}lllcc>{\columncolor[HTML]{D6D3D2}}cccccc>{\columncolor[HTML]{D6D3D2}}c>{\columncolor[HTML]{C8BBFC}}c@{}}
\toprule
Prompt &Query  & Answer & \multicolumn{3}{c|}{Ans. to Ques. (\AQ)} & \multicolumn{6}{c|}{Ans. to Context (\AC)}& \cellcolor[HTML]{FFFFFF}\\ 
\cmidrule{4-13}

Setting  & LLM & LLM & AlignSc. & Relev. &Avg. & AlignSc. & ROUGE-1 & BERTSc. & FactCC & Faith.  & Avg. & Avg. \\\midrule
\SQ     & GPT3.5  & GPT4  & 0.678 & 0.835 & \textbf{0.757} & 0.655 & 0.006 & 0.796 & 0.651 & 0.080 & 0.438 & 0.529 \\
       & GPT4    & GPT4  & 0.688 & 0.798 & 0.743 & 0.671 & 0.006 & 0.797 & 0.633 & 0.091 & 0.440 & 0.526 \\
       & GEMMA   & GPT4  & 0.688 & 0.801 & 0.745 & 0.671 & 0.006 & 0.797 & 0.633 & 0.091 & 0.440 & 0.527 \\
       & Mistral & GPT4  & 0.666 & 0.783 & 0.725 & 0.667 & 0.009 & 0.797 & 0.416 & 0.035 & 0.385 & 0.482 \\\cmidrule{2-13}
       & GPT3.5  & QWEN & 0.687 & 0.816 & \underline{0.752} & 0.593 & 0.004 & 0.788 & 0.592 & 0.089 & 0.413 & 0.510 \\
       & GPT4    & QWEN & 0.654 & 0.818 & 0.736 & 0.601 & 0.005 & 0.792 & \textbf{0.656} & 0.072 & 0.425 & 0.514 \\
       & GEMMA   & QWEN & 0.654 & 0.820 & 0.737 & 0.601 & 0.005 & 0.792 & 0.655 & 0.072 & 0.425 & 0.514 \\
       & Mistral & QWEN & 0.659 & 0.813 & 0.736 & 0.576 & 0.008 & 0.792 & 0.562 & 0.052 & 0.398 & 0.495 \\\midrule
\MQCAT  & GPT4    & GPT4  & 0.620 & 0.788 & 0.704 & 0.765 & 0.043 & \textbf{0.874} & 0.458 & 0.396 & 0.507 & \underline{0.563} \\
       & Mistral & GPT4  & 0.619 & 0.796 & 0.708 & 0.753 & 0.040 & 0.870 & 0.509 & 0.396 & \textbf{0.514} & \textbf{0.569} \\\cmidrule{2-13}
       & GPT4    & QWEN & 0.649 & 0.724 & 0.687 & 0.651 & \textbf{0.050} & 0.841 & 0.294 & 0.464 & 0.460 & 0.525 \\
       & Mistral & QWEN & 0.592 & 0.722 & 0.657 & 0.680 & 0.036 & 0.849 & 0.512 & 0.483 & \underline{0.512} & 0.553 \\\midrule
\MQSTCK & GPT4    & GPT4  & 0.687 & 0.811 & 0.749 & \textbf{0.787} & 0.039 & 0.868 & 0.231 & \textbf{0.609} & 0.507 & 0.576 \\
       & Mistral & GPT4  & 0.615 & 0.789 & 0.702 & 0.759 & 0.040 & 0.863 & 0.372 & 0.480 & 0.503 & 0.560 \\\cmidrule{2-13}
       & GPT4    & QWEN & 0.621 & 0.734 & 0.678 & 0.686 & 0.038 & 0.845 & 0.107 & 0.489 & 0.433 & 0.503 \\
       & Mistral & QWEN & 0.617 & 0.811 & 0.714 & 0.683 & 0.042 & 0.845 & 0.114 & 0.483 & 0.433 & 0.514 \\
\bottomrule
\end{tabular}%
}
\caption{\OzWheat metric scores evaluating the answer quality. Answers are generated by GPT4 and QWEN.}
\label{tab:ozwheat_ans_metrics}
\end{table*}

\section{Automatic Metric Results}
\label{sec:app_metric_results}

We found that there is no single language model that performs well across all stages of the RAG pipeline. As shown in the results in Table~\ref{tab:frosttolerance_metrics} on the \FrostTolerance dataset, the highest average metric scores were achieved by different models for different evaluation dimensions. For example, for the query-to-question (\QrQ) quality, GPT4 generated queries in \MQCAT had the best scores. However, GPT4 in \MQSTCK setting generated the best contexts aligned to the question (context-to-question, \CQ) and GPT4 in \MQCAT generated the best contexts with respect to the retrieved abstracts (context-to-abstract, \CAb). 

For the answer quality (Table~\ref{tab:frosttolerance_ans_metrics}), none of the contexts that were generated by the better quality LLMs were able to produce better answers aligned to the question (answer-to-question, \AC) or context (answer-to-context, \AC). Looking at the summary of the best performing LMs in different settings from Table~\ref{tab:metric_summary}, we observe a similar trend across all datasets. These results essentially indicate that rather than depending on one LLM at each stage of the pipeline, it is better to select a combination of LLMs through multiple evaluation dimensions for optimal outcome. 

Furthermore, we present the detailed individual metrics scores and averaged scores on each dimension in Table~\ref{tab:frosttolerance_metrics}-\ref{tab:frosttolerance_ans_metrics} on \FrostTolerance and Table~\ref{tab:ozwheat_metrics}-\ref{tab:ozwheat_ans_metrics} on \OzWheat datasets. 
Looking at these detailed scores 
we observe that, because different evaluation metrics emphasise different quality aspects of the evaluated texts, no single metric provides a complete picture. 
For example, from the FactCC and Faithfulness metric scores for \FrostTolerance and \OzWheat datasets in the Table~\ref{tab:frosttolerance_ans_metrics} and \ref{tab:ozwheat_ans_metrics}, we observe that the answers generated by using the contexts in the single query setting are factually more accurate, but less faithful than the others. 
A combination of metrics is therefore more effective, with each metric contributing differently to the assessment of LLM response quality.

\begin{figure}[ht!]
    \centering
    \begin{subfigure}{\columnwidth} 
    \centering
        \includegraphics[width=\linewidth]{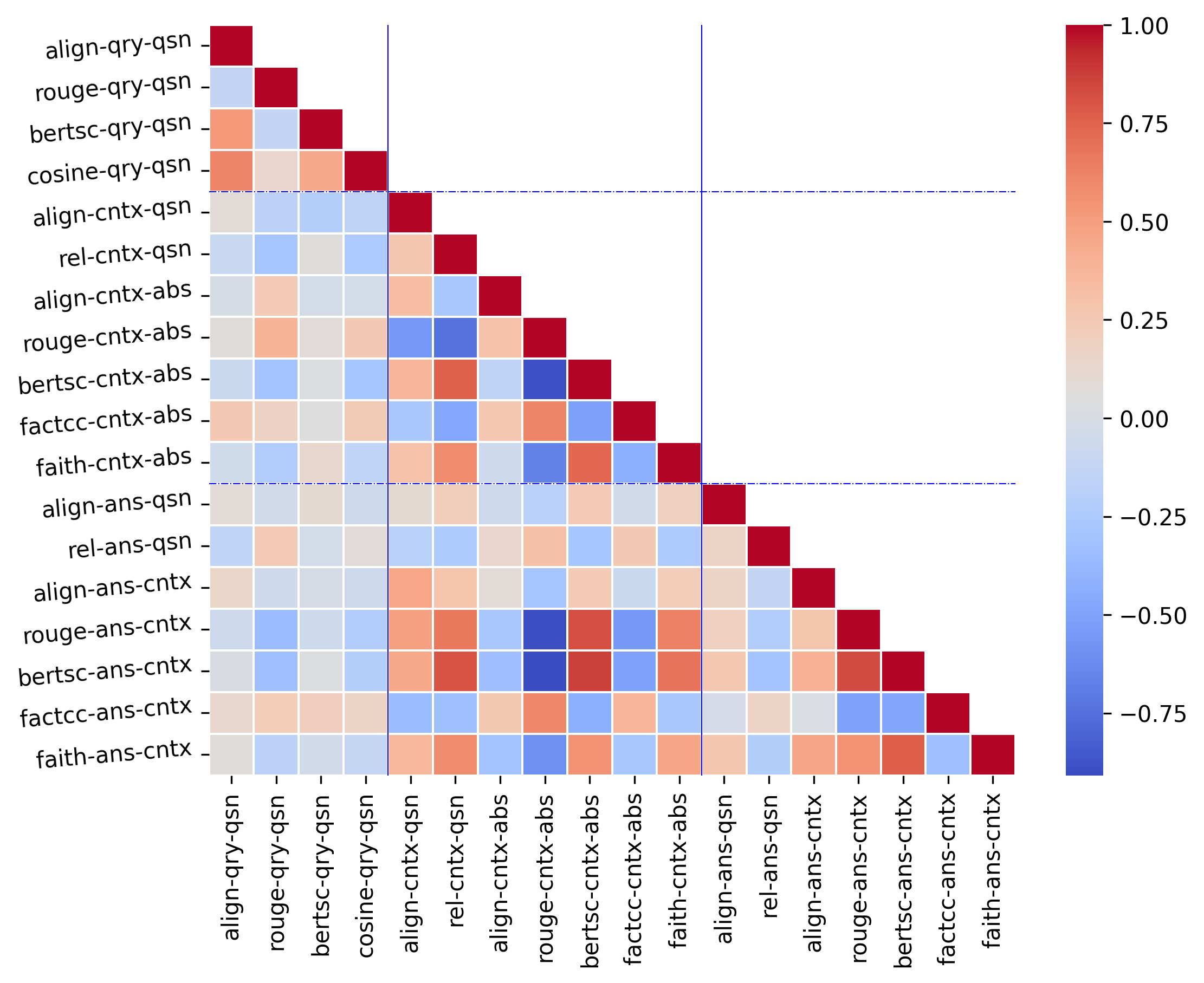} 
        \caption{}
        \label{fig:sub_FT_corr}
    \end{subfigure}%
    \hspace{0.5cm} 
    \begin{subfigure}{\columnwidth}
        \centering
        \includegraphics[width=\linewidth]{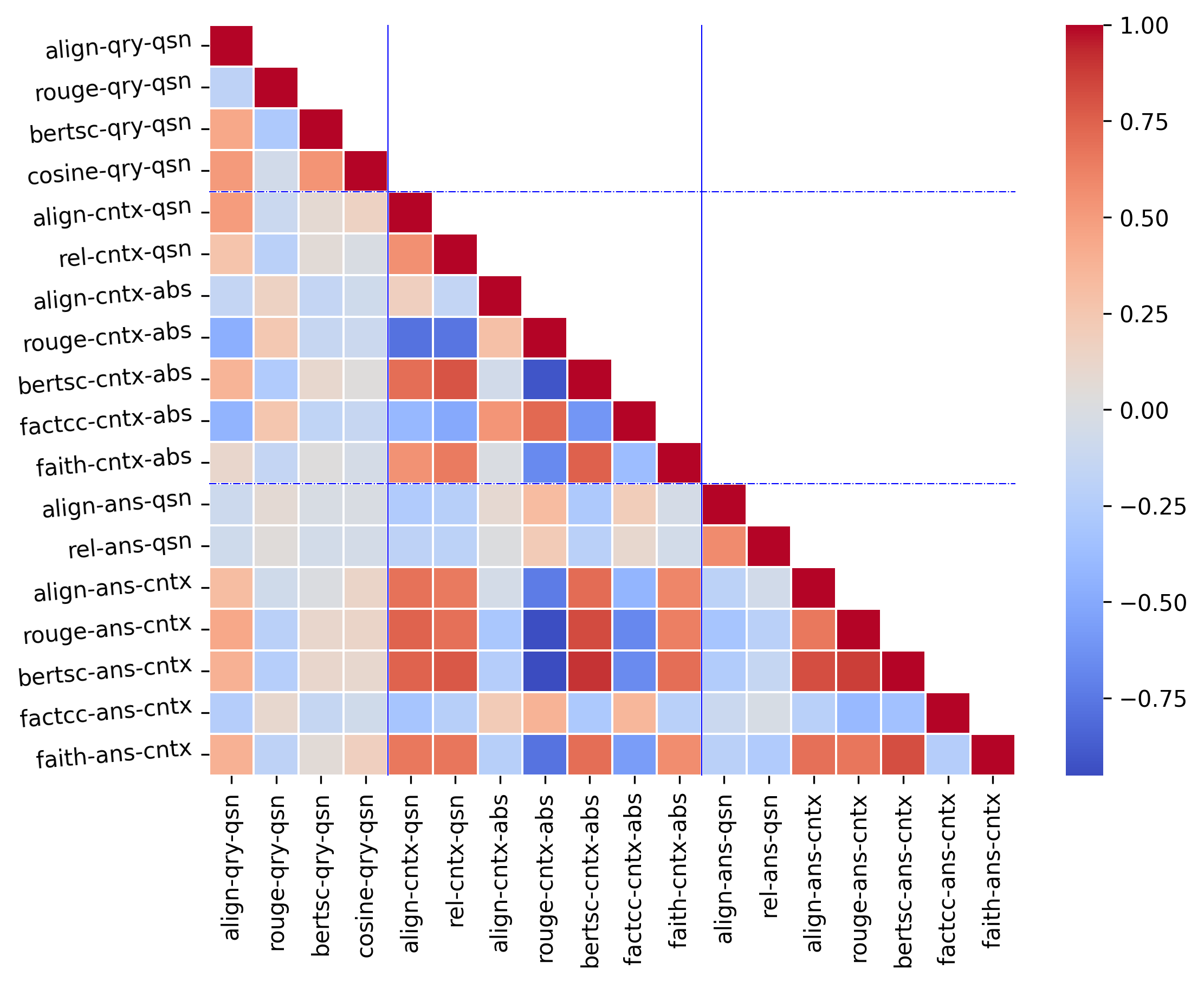}
        \caption{}
        \label{fig:sub_OzW_corr}
    \end{subfigure}%
    \caption{Pearson's correlation coefficients ($r$) between the evaluation metrics for a) \FrostTolerance b) \OzWheat datasets.}
    \label{fig:metric_metric_corr}
\end{figure}

Through an analysis of the Pearson correlation coefficient $r$ across different metric combinations, we observe varying levels of correlation between evaluation metrics, as illustrated in the Figure~\ref{fig:metric_metric_corr}.
Across all datasets, metrics assessing answer–context alignment (the \AC dimension) exhibit relatively higher correlation with metrics measuring context–question alignment (the \CQ dimension). In contrast, query–question evaluation metrics (\QrQ) show consistently lower correlation with metrics from other evaluation dimensions. Beyond these trends, no consistent correlation patterns are observed across the remaining metric combinations. This analysis indicates that there is no one-size-fits-all metric configuration for evaluating RAG systems, and it is crucial to incorporate human guidance when selecting and combining evaluation metrics.

In Table~\ref{tab:annot_metric_agreement_rates}, we report the agreement rates between uniform-weighted metric-based preferences and human preferences (human annotated set) and LLM preferences (synthetic annotations).

\begin{table}
\small
\centering
\resizebox{\columnwidth}{!}{%
\begin{tabular}{@{}llcccc@{}}
\toprule
Metric       & Hum-Metric &Pearson's & LLM-Metric &Pearson's  \\
             & Agree. (\%)& $r$ & Agree. (\%)& $r$ \\\midrule
All          & 61.45   & \textbf{0.300 (p=0.002)} & 47.163 & \textbf{0.267 (p=0.005)} \\\midrule
\QrQ &&&&\\\midrule
AlignScore   & 59.04 &  0.164 (p=0.035)    & 48.27 & -0.044 (p=0.005)  \\
ROUGE        & 52.41 &  -0.010 (p=0.902)    & 46.99& -0.069 (p<0.001)  \\
BERTScore    & 50.60 &  -0.003 (p=0.974)    & 46.77& -0.079 (p<0.001)  \\
CosSim.      & 51.20  & 0.002 (p=0.976) & 47.85 & -0.050 (p=0.001) \\\midrule
\CQ &&&&\\\midrule
AlignScore   & 56.63 & 0.094 (p=0.228)  & 64.37 & \textbf{0.262 (p<0.001)}    \\
Relevance    & 51.81   & -0.008 (p=0.919) & 55.75& \textbf{0.082 (p<0.001)} \\\midrule
\CAb &&&&\\\midrule
AlignScore   & 53.01 & 0.032 (p=0.679)   & 49.23 &  -0.023 (p=0.138)   \\
ROUGE        & 49.40   & -0.046 (p=0.557) & 38.82&  -0.226 (p<0.001)   \\
BERTScore    & 50.60   & -0.021 (p=0.790) & 54.72& 0.055 (p = 0.003)    \\
FactCC       & 61.45   & \textbf{0.298 (p=0.010)} & 42.77& -0.138 (p<0.001)   \\
Faithfulness & 50.60   & -0.018 (p=0.822) &57.18 & 0.106 (p<0.001) \\\midrule
\AQ &&&&\\\midrule
AlignScore   & 61.64 & \textbf{0.301 (p=0.050)} & 56.42&  0.124 (p<0.001)   \\
Relevance    & 39.73  &  {-0.230 (p=0.051)} & 43.02&  -0.127 (p<0.001)  \\\midrule
\AC &&&&\\\midrule
AlignScore   & 65.75 &  \textbf{0.313 (p=0.007)} & 68.00&  \textbf{0.370 (p<0.001)}  \\
ROUGE        & 52.05   &    0.245 (p=0.704) & 58.85& 0.143 (p<0.001)\\
BERTScore    & 57.53   &   0.253 (p=0.197)  & 68.20& \textbf{0.354 (p<0.001)}\\
FactCC       & 57.53   &   0.245 (p=0.219)  & 40.83& -0.148 (p<0.001) \\
Faithfulness &  61.64   &   \textbf{0.303 (p=0.039)}  & 61.82& \textbf{0.208 (p<0.001)} \\ \bottomrule
\end{tabular}%
}
\caption{Rates of agreements between uniform-weighted metric-based preferences against human preferences (human annotated set) and LLM preferences (synthetic annotations). Bold indicates statistical significance.}
\label{tab:annot_metric_agreement_rates}
\end{table}

\begin{table}[t!]
\centering
\resizebox{\columnwidth}{!}{%
\begin{tabular}{@{}p{10cm}@{}}
\toprule
Rubric: Generic                                   \\ \midrule
- Does the response answer the user's question?                                      \\
- Is the answer easily understandable and complete with non-repetitive information?  \\
- Are the references useful to answer the question?                                  \\
- Does the answer cite the references correctly for each claim?                     \\
- Can the claims in the answers be accurately supported by the given references?     \\\midrule
Rubric: Domain-specific \\\midrule
- Does the answer include a short description of the function of the candidate gene? \\
- Does the answer state if it can or cannot find evidence linking the candidate gene to the target trait? \\
- If evidence has been identified, are the type or types of evidence described?      \\
- Does the answer reference the source of all evidence linking the candidate gene to the target trait?    \\ \bottomrule
\end{tabular}%
}
\caption{Used generic and domain-specific rubrics to guide the LLM in judging the pairwise answers}
\label{tab:app_llm_rubrics}
\end{table}

\section{LLM-as-a-Judge}
\label{sec:app_llm_judge}

Following the work of \citet{lin2025wildbench}, we create in-prompt rubrics containing questions to check the quality of the answers as a guideline for the LLM to generate preferences over pairwise answers. The used rubric questions are listed in Table~\ref{tab:app_llm_rubrics}. The generic rubric is created by taking motivation from WildBench \citep{lin2025wildbench} and LLM-rubric \citep{hashemi-2024-llm} and connecting with the evaluation dimension explored in our work. To create the domain-specific rubric, we consulted the domain experts.

\section{Quality Dimensions vs. human reasonings}
\label{sec:app_quality_dim_analysis}

We prompt \textit{GPT4o-mini} in a one-shot prompt setting to group similar human reasonings and label the quality aspects considered in the reasons. Table~\ref{tab:app_prompt_gen_dimension} shows the used prompt for this purpose. Some example annotations are illustrated in Tables~\ref{tab:app_annot_exmpl1}\&\ref{tab:app_annot_exmpl2}. The resultant criteria and the human reasoning under each criterion are presented in Table~\ref{tab:app_gpt4_dimensions}. Based on the criteria labels, the frequency of each criterion in the case of mismatched human-metric criteria for `all' and a subset of metrics is in Table~\ref{tab:app_dim_mismatched}. 

\begin{table}[t!]
\small

    \begin{tabular}[width=\columnwidth]{p{0.9\columnwidth}}
    \cellcolor{black}\textbf{\textcolor{white}{Prompt: Grouping Human Ratoinals}}
    \\\midrule
    human reasonings:
    \{reasons\}
    \\
    \\
    
    Given the above list of reasons for not preferring some LLM-generated answers, try to group them into reasoning categories.
    \\
    \\
    Give a short category name for each category.    
    \\
    \\
    Return the output in a JSON firmat, where "criteria" is the "<category label>" and "reasons" is the list of reasons fall into the category.
    \\
    For example, "criteria": "answer relevance", "reasons": ['The answer is not as relevant as it also does not provide specific information regarding the proportion of studies linking Ethylene-responsive transcription factors with cold acclimation.',
    'The answer does not provide any relevant supporting evidence that links ethylene-responsive transcription factors to cold acclimation, making it less relevant.']
    \\
    Your response:\\
    \bottomrule
    \end{tabular}
    \caption{One-shot LLM prompt to group the human reasonings for not preferring an answer and create quality dimension.}
    \label{tab:app_prompt_gen_dimension}
\end{table}


\section{SciArena Dataset}
\label{sec:app_sciarena}

The \SciArena \citep{zhao2025sciarena} is an open, crowdsourced benchmarking platform designed to evaluate foundation models on scientific literature-grounded tasks. The dataset\footnote{\url{https://huggingface.co/datasets/yale-nlp/SciArena}} contains 13,204 human preferences over pairwise LLM-generated answers on 25 subjects. From this dataset, we investigate metric weight calibration on the subjects closer to our focus domain. The list of subjects and their data size is reported in Table~\ref{tab:sciarena_data} and an example question and answer pair is given in Table~\ref{tab:sciarena_exmpl}. Note, we excluded the `tie' or `bad' annotations from the investigation.

\begin{table}[t!]
\centering
\resizebox{0.5\columnwidth}{!}{%
\begin{tabular}{lc}
\toprule
 Subject & Datasize\\\midrule
Combined &   2251  \\\midrule
Biomed. &   662  \\
Medicine&   801  \\
Neuroscience& 206 \\
Pharmacy&   334  \\
Public Health& 248 \\\bottomrule
\end{tabular}%
}
\caption{Dataset size of the different subjects in \SciArena}
\label{tab:sciarena_data}
\end{table}

\begin{table}[t!]
\centering
\resizebox{\columnwidth}{!}{%
\begin{tabular}{llllllllll}
\toprule
 &
  Total &
  \rotatebox{90}{Similarity to the   Selected Answer} &
  \rotatebox{90}{Relevance of Answer} &
  \rotatebox{90}{Lack of Details} &
  \rotatebox{90}{Lack of Supporting Evidence} &
  \rotatebox{90}{Incorrect or irrelevant context} &
  \rotatebox{90}{Organization and Structure} &
  \rotatebox{90}{Answer Clarity and Fluency} &
  \rotatebox{90}{Inaccurate Claims} \\\midrule
All      & 54 & 16 & 15 & 9  & 4 & 4 & 3 & 2 & 2 \\\midrule
Qr2Q     & 70 & 15 & 28 & 11 & 7 & 4 & 1 & - & 1 \\\midrule
C2Q      & 62 & 18 & 22 & 10 & 6 & 3 & 2 & - & 1 \\
C2Ab     & 75 & 17 & 30 & 12 & 7 & 6 & 1 & 1 & 1 \\
C2Q+C2Ab & 70 & 18 & 27 & 11 & 6 & 6 & - & 1 & 1 \\\midrule
A2Q      & 72 & 21 & 26 & 8  & 6 & 5 & 3 & 1 & 2 \\
A2C      & 56 & 19 & 15 & 8  & 5 & 4 & 2 & 2 & 1 \\
A2Q+A2C  & 57 & 19 & 14 & 9  & 4 & 5 & 3 & 2 & 1 \\\bottomrule
\end{tabular}%
}
\caption{Criteria from human reasons grouped by LLM for mismatched human-metric preference cases.}
\label{tab:app_dim_mismatched}
\end{table}

\section{Human Annotation App}
\label{sec:app_human_annot_app}
The annotation application was hosted on an internal server. We provided secure logins to the annotators to access the survey application. We presented both answer pairs on a single screen for each question. One answer in the second pair is loaded based on the choice of the first answer. 
The reasons for not preferring an answer are collected as free texts. Also, we employed a separate text box for the annotators to provide stylistic reasons. For ease of reading, we highlighted the same PubMed IDs (PMIDs) with one color.

\begin{table*}[t!]
\centering
\resizebox{\textwidth}{!}{%
\begin{tabular}{@{}lcp{0.7\textwidth}@{}}
\toprule
Criteria &
 Frequency & Example Reasons \\ \midrule
Relevance of Answer & 54 &`the answer is not as relevant', `answer is not relevant to the question',`the answer is neither sufficiently relevant nor complete.', `not as relevant to the candidate gene', `the answer is not as relevant and does not provide inline citations', `the answer is very similar to the selected answer but less relevant', `not as relevant does not get to answer', `not as relevant, no description of gene function', `the answer is neither sufficiently relevant nor complete.'
\\\midrule
Similarity to the Selected Answer &35 & `the answer is very similar to the selected answer but not as concise.', `the answer has less information about the other reference.', `similar, irrelevant context', `both bad, no description of gene function', `the answer is almost similar to the selected answer' \\\midrule
Lack of details & 16 & `the answer is not as detailed. The paragraph structure is not helpful for directly answer the question.', `the answer has less information', `the answer is not as detailed.', `the answer has less information.', `details provided could be misleading to a novice.', `answer is not as detailed', `less detail', `the answer is neither sufficiently relevant nor com plete.', `the answer is not complete' \\\midrule
Incorrect or Irrelevant Context & 11 & `incorrect context', `irrelevant context, no description of gene function', `context has low relevance'\\\midrule
Lack of Supporting Evidence & 11 & `no supporting evidence', `the answer does not provide inline citations.', `the answer is not as relevant and does not provide inline citations', `did not define function of gene', `irrelevant; did not define function of gene', `not as relevant, no description of gene function', `part of the answer is not as relevant and there is no citations', `the answer is not as relevant, no supporting evidence' \\\midrule

Organization and Structure & 7 & `the paragraph structure is not helpful for directly answer the question.', `the paragraph structure and organization are not as useful', `the answer is not as relevant paragraph structure and organization are not helpful (not using topic sentence structure)', `incomplete, intro too long', `The phrasing in the answer is not as fluent.'
\\ \midrule
Answer Clarity and Fluency & 4 & `The phrasing in the answer is not as fluent.', `Answer did not properly address the two parts of the question.', `Answer did not fully address the question.', `The answer is very similar to the selected answer but has less information.The phrasing in the answer is not as fluent.'\\\midrule
Inaccurate Claims & 2 & `Answer has incorrect claims / evidence', `not as accurate' \\\midrule
\textbf{Total reasons}& \textbf{140} & \\
\bottomrule
\end{tabular}%
}
\caption{LLM generated quality dimensions from the human reasons. }
\label{tab:app_gpt4_dimensions}
\end{table*}

\begin{table*}[]
\centering
\begin{tabular}{@{}p{15cm}@{}}
\toprule
\textbf{Example \#1} \\ \midrule
\textbf{Question:} What proportion of Transcriptomic studies on plants found Gpcr-type g protein 2 gene to be associated with cold acclimation? \\\midrule
\textbf{Human reason:} The answer has less information \\\midrule
\textbf{Human preference:} Answer 2  \textbf{Metric preference:} Answer 1 \\\midrule
\textbf{Answer 1:} "The provided context does not mention any specific transcriptomic studies on plants that associate the GPCR-type G protein 2 gene with cold acclimation. The cited studies focus on viroid effects on gene expression (Joubert 2022 - 35887025), stress responses via RNA-seq and single-cell RNA-seq (Rurek 2024 - 39524930), and single-cell transcriptomics for plant development (Yu 2023 - 36540021). None of these contexts address GPCR-type G protein 2 gene functionality or its relationship with cold acclimation. Therefore, the proportion cannot be determined.", \\\midrule
\textbf{Anwer 2:} "The provided context does not specify the proportion of transcriptomic studies on plants that found the G-protein coupled receptor (GPCR) type G protein 2 gene to be associated with cold acclimation. It emphasizes the general mechanisms of cold acclimation, such as physiological adjustments and changes in gene expression involving C-repeat binding factors (CBFs) (Barrero-Gil 2019 - 30288701), as well as the role of various proteins in regulating this process (Barrero-Gil 2013 - 23498862). However, specific information regarding the GPCR-type G protein 2 gene's association with cold acclimation is not included in the provided studies. Thus, I cannot provide a definitive answer regarding the proportion." \\\midrule
\textbf{Example \#2 }\\\midrule
\textbf{Question:} Candidate gene TraesCS6B03G1296700 on genome GCF\_018294505.1 has the annotation of being a SKIP interacting protein 16 gene. Are there any known associations or links between this family of genes and the trait height in wheat research? Please also describe the evidence for the association. \\\midrule
\textbf{Human preference:} Answer 1 \textbf{Metric preference:} answer 2 \\\midrule
\textbf{Answer 1: }"The context does not provide direct evidence linking the SKIP interacting protein 16 gene (TraesCS6B03G1296700) to wheat height. While SKIP is described as a conserved splicing factor and transcriptional regulator in plants (Li 2022 - 35640153), its role in wheat or specific associations with plant height are not mentioned. The TB1 gene in wheat is noted to influence plant height through allelic variation (Dixon 2021 - 32449515), but this is unrelated to SKIP. Additionally, SKIP's role in inhibiting sphingosine kinase (Ghazaly 2020 - 32161116)." \\\midrule
\textbf{Answer 2: }"There is limited direct evidence linking SKIP interacting protein 16 genes, including TraesCS6B03G1296700, to height traits in wheat. However, studies on plant height as an important agronomic trait highlight the role of various genes in influencing this characteristic. The Reduced height-B1b (Rht-B1b) and Rht-D1b alleles have been specifically associated with semi-dwarf plant architecture, enhancing lodging resistance (Song 2023 - 37100915). Furthermore, QTL analyses have been effective in identifying and estimating the effects of genes related to plant height (Paccapelo 2022 - 35597886; Zhou 2023 - 37349717). While the specific role of SKIP interacting protein 16 in height traits remains to be elucidated, its interaction in gene networks may warrant further investigation in the context of height-related QTLs." \\ \bottomrule
\end{tabular}%
\caption{Example of human annotation and metric-based preference mismatch when two answers are very similar.}
\label{tab:app_annot_exmpl1}
\end{table*}

\begin{table*}[t!]
\centering
\begin{tabular}{@{}p{7.5cm}|p{7cm}@{}}
\toprule
\multicolumn{2}{p{14cm}}{\textbf{Question:}}\\\midrule
\multicolumn{2}{p{14cm}}{Candidate gene TraesCS6B03G0835400 on genome GCF\_018294505.1 has the annotation of being a SBP (S-ribonuclease-binding protein) family protein gene. Are there any known associations or links between this family of genes and the trait spike length in wheat research? Please also describe the evidence for the association.} \\ \midrule
\multicolumn{2}{p{15cm}}{\textbf{Answer 1:}} \\\midrule
\multicolumn{2}{p{15cm}}{The SBP (S-ribonuclease-binding protein) family gene TraesCS6B03G0835400 is linked to spike length in wheat. Research indicates that SBP genes, such as BdSBP1 and BdSBP3, play roles in spike development, with their expression patterns during spikelet initiation and responses to environmental factors like vernalization and temperature (Tripathi 2020 - 32929136). A T-DNA mutant of Bdsbp9 showed reduced spike length, highlighting the functional importance of SBP genes in spike development (Tripathi 2020 - 32929136).} \\\midrule
\multicolumn{2}{p{15cm}}{\textbf{Answer 2:}}\\\midrule
\multicolumn{2}{p{15cm}}{There is currently no direct evidence linking the SBP (S-ribonuclease-binding protein) family genes, such as TraesCS6B03G0835400, to spike length traits in wheat research based on the provided context. The context discusses the association of specific haplotypes in different genes with spike length traits (Zhang 2023 - 37310852) but does not mention SBP family genes or their role in determining spike length. Additionally, while SPL genes are mentioned in relation to plant growth traits (Zhao 2022 - 36011366), they do not directly address the SBP family or spike length in wheat. Therefore, further research would be necessary to explore any potential associations between SBP genes and spike length in wheat.} \\
\midrule
\midrule
Annot. 1&
  Annot. 2\\
  \midrule
Preference: \#1 &
Preference: \#1 \\
Rationale: &
Rationale:\\ 
Incorrect claim or evidence &
The answer is not as relevant \\
\bottomrule
\end{tabular}%
\caption{Different reasonings for the same answer pair and the same preferences by two annotators.}
\label{tab:app_annot_exmpl2}
\end{table*}

\begin{table*}[t!]
\centering
\begin{tabular}{@{}p{16cm}@{}}
\toprule
\textbf{Subject:} Biomed. Engi.\\ \midrule
\textbf{Question:} What recent progress characterises novel monolithic stationary phases for liquid chromatography? \\\midrule
\textbf{Llama-4-Scout: }The development of novel monolithic stationary phases has significantly advanced liquid chromatography, offering improved separation efficiency, selectivity, and versatility. Recent progress in this area includes the introduction of nanomaterial-based stationary phases, such as nanoporous silica, zirconia, and graphene, which provide ultra-high surface areas and nanoscale confinement effects (Ratnam et al. 2023b).\\
Monolithic columns, made from organic polymers, inorganic materials, or hybrid structures, have been engineered to offer tailored characteristics for specific analyses (Nechv\u00e1talov\u00e1 et al. 2022a). These columns have demonstrated superior permeability, efficiency, and ease of preparation, making them attractive alternatives to traditional particle-packed columns (Gregu\u0161 et al. 2023a). The incorporation of metal ions, such as Co(II), into molecularly imprinted polymers has enhanced selectivity for separating enantiomeric compounds (Amalia et al. 2022b). Additionally, zwitterionic sulfobetaine-based monoliths have been developed for hydrophilic interaction chromatography, exhibiting good selectivity and efficiency (Wang et al. 2022a).\\
The use of bio-inspired materials, like lipase-immobilized monoliths, has also shown potential for chiral separations (Fouad et al. 2023a). Furthermore, the integration of monolithic columns with mass spectrometry has enabled sensitive and high-throughput analysis of complex samples (Shamsi et al. 2022a). Overall, these advancements in monolithic stationary phases are transforming liquid chromatography, enabling faster, more sensitive, and high-resolution separations for a wide range of applications (Ratnam et al. 2023b).\\\midrule
\textbf{Deepseek-R1:} Recent advancements in monolithic stationary phases for liquid chromatography (LC) emphasize enhanced selectivity, fabrication techniques, and application-specific designs. Gradient monolithic phases, achieved through controlled chemical modifications like the controlled rate infusion method or photo-grafting, enable gradual changes in surface functionality along the column, improving resolution for complex biological samples (Alla et al. 2022a). Metal-mediated molecularly imprinted polymers (MMIPs) have emerged for chiral separations, with cobalt(II)-based monolithic columns demonstrating high enantioselectivity for racemic citronellal, leveraging metal coordination to enhance template recognition (Amalia et al. 2022b).\\
Hybrid organic-inorganic monoliths, such as polymer-bound silica phases, combine high permeability with exceptional efficiency (~288,000 plates/m for peptides), addressing challenges in biomolecule separation (Ali et al. 2024b). Innovations in miniaturization include microbore columns (1 mm i.d.) that reduce solvent consumption while maintaining performance, ideal for environmentally friendly LC systems (Amalia et al. 2022a). Zwitterionic sulfobetaine-based monoliths, fabricated via rapid photoinitiated copolymerization, offer superior hydrophilic interaction chromatography (HILIC) capabilities, benefiting polar analyte separations (Wang et al. 2022a).\\
Composite monoliths integrating nanomaterials (e.g., carbon nanotubes) or metal-organic frameworks (MOFs) enhance surface area and selectivity, particularly for proteomic and pharmaceutical analyses (Nechv\u00e1talov\u00e1 et al. 2022a). Additionally, advancements in nano-LC utilize silica-based monolithic columns with sub-1 \u00b5m pores for sensitive food analysis, achieving low back-pressure and high mass transfer kinetics (Aydo\u011fan et al. 2023a). These developments highlight trends toward multifunctional, application-tailored monoliths, improved fabrication reproducibility, and integration with miniaturized systems, expanding their utility in biomedicine, environmental monitoring, and industrial chromatography (Gregu\u0161 et al. 2023a, Ratnam et al. 2023c).\\\midrule
Human preference: 1                                                       \\ \bottomrule
\end{tabular}%
\caption{Example questions and answers from \SciArena dataset for Biomedical Engineering domains.}
\label{tab:sciarena_exmpl}
\end{table*}

\end{document}